\documentclass{article}

\usepackage{geometry}
 \usepackage{xcolor}         
\usepackage[utf8]{inputenc}
\usepackage[T1]{fontenc}
\usepackage{microtype}
\usepackage{graphicx}
\usepackage{subcaption}
\usepackage{booktabs} 

\usepackage{amsmath}
\usepackage{amssymb}
\usepackage{mathtools}
\usepackage{amsthm}

\usepackage{amssymb}
\usepackage{dirtytalk}
\usepackage{float}

\usepackage{multicol}
\usepackage{authblk}
\usepackage{wrapfig}
\usepackage{stmaryrd}

\newcommand{\checked}[1]{#1}

\usepackage[textsize=tiny]{todonotes}

\usepackage{algorithm}
\usepackage{algpseudocode}
\usepackage{amsmath,amsthm,mathtools}

\newcommand{\dataset}{\mathcal{D}}

\title{Continual Feature Selection: Spurious Features in Continual Learning}

\author{%
  Timothée Lesort \\
  Université de Montréal, MILA-Quebec AI Institute\\
  \texttt{timothee.lesort at umontreal.ca} \\

}

\begin{document}

\maketitle

\begin{abstract}

Continual Learning (CL) is the research field addressing learning without forgetting when the data distribution is not static. 
This paper studies spurious features' influence on continual learning algorithms.
We show that continual learning algorithms solve tasks by selecting features that are not generalizable. 
Our experiments highlight that continual learning algorithms face two related problems: (1) spurious features and (2) local spurious features. The first one is due to a covariate shift between training and testing data, while the second is due to the limited access to data at each training step.
We study (1) through a consistent set of continual learning experiments varying spurious correlation amount and data distribution support.
We show that (2) is a major cause of performance decrease in continual learning along with catastrophic forgetting. 
This paper presents a different way of understanding performance decrease in continual learning by highlighting the influence of (local) spurious features in algorithms capabilities.

\end{abstract}

\section{Introduction}

Feature selection is a standard machine learning problem. Its objective is to improve the prediction performance, provide faster and more effective predictors, and provide a better understanding of the underlying process that generated the data \cite{Guyon2003Introduction}. In this paper, we are interested in improving prediction performance in the presence of spurious features. Spurious features arise when features correlate well with labels in training data but not in test data. Learning algorithms that rely on spurious features will generalize badly to test data. 

In continual learning (CL), the training data distribution changes through time. %
Hence, we could expect that spurious features (SFs) in one time-step of the data distribution will not last. A continual learning algorithm that selects a spurious feature to solve a task can then be resilient and learn better features later, given more data. 
Algorithms can aim to detect and ignore spurious features learned in the past \cite{javed2020learning}. 
An example of a task with spurious features could be a classification task between cars and bikes, but in the training data, all cars are red, and all bikes are white. A model could easily overfit the color feature to solve the task while it is not discriminative in the test data. This problem is notably caused by a covariate shift between train data and test data. In CL, we expect that future tasks will bring pictures of different cars and bikes that will allow the model to learn better features.

On the other hand, in CL, a second type of spurious feature can be described: \textit{local spurious features}. Local features denote features that correlate well with labels within a task (a state of the data distribution) but not in the full scenario. 
In opposite to usual \textit{spurious features}, this problem is provoked by the unavailability of all data, for example because only red cars and white bikes are currently available in train data, and is not by a covariate shift between train and test. It is, therefore, a problem specific to continual learning.

\begin{figure*}
    \centering
    \begin{subfigure}[b]{0.24\linewidth}
        \centering
        \includegraphics[width=\linewidth]{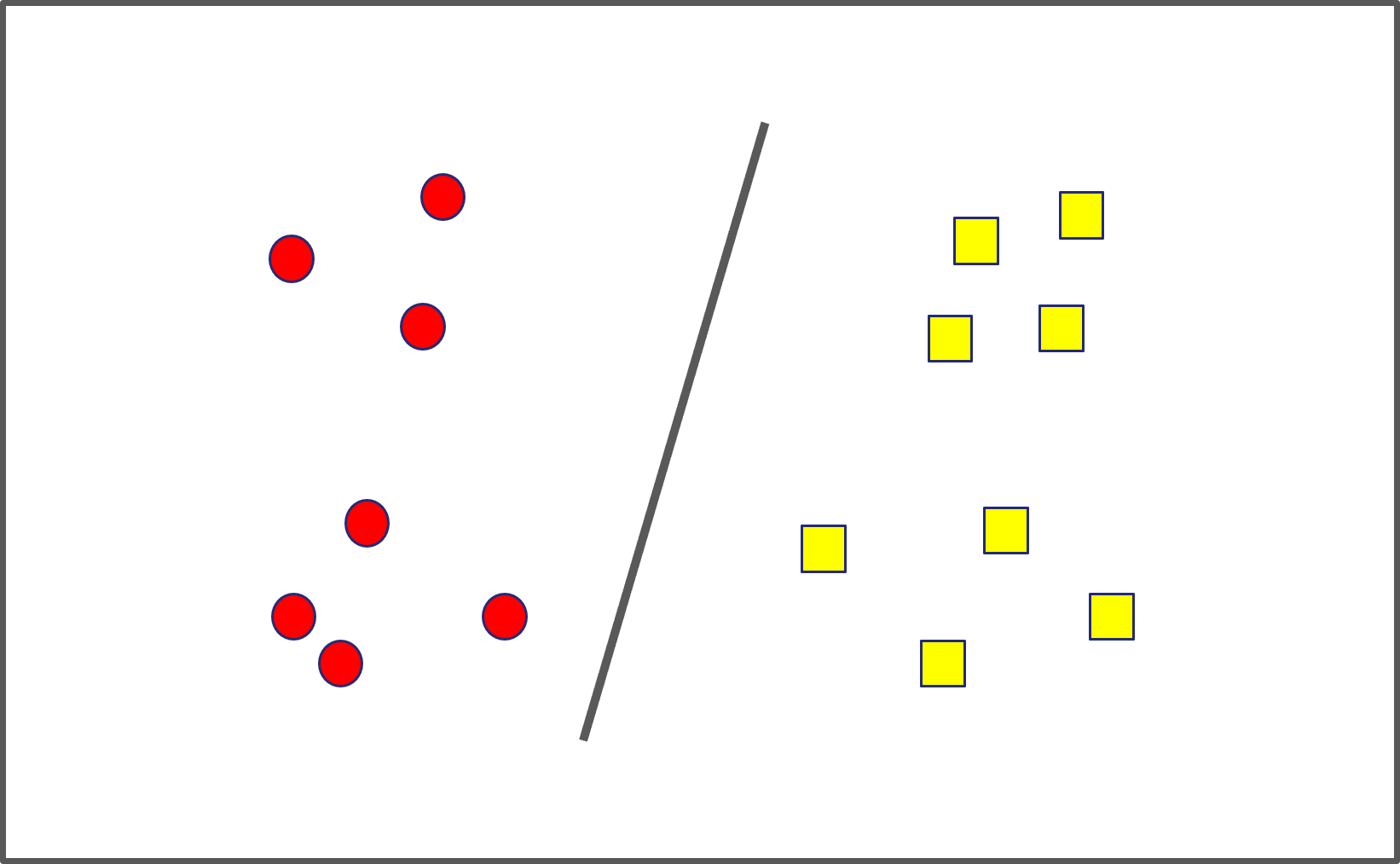}
    \caption{Train (Spurious)}
    \label{fig:spurious_train}
    \end{subfigure}
    \begin{subfigure}[b]{0.24\linewidth}
        \centering
        \includegraphics[width=\linewidth]{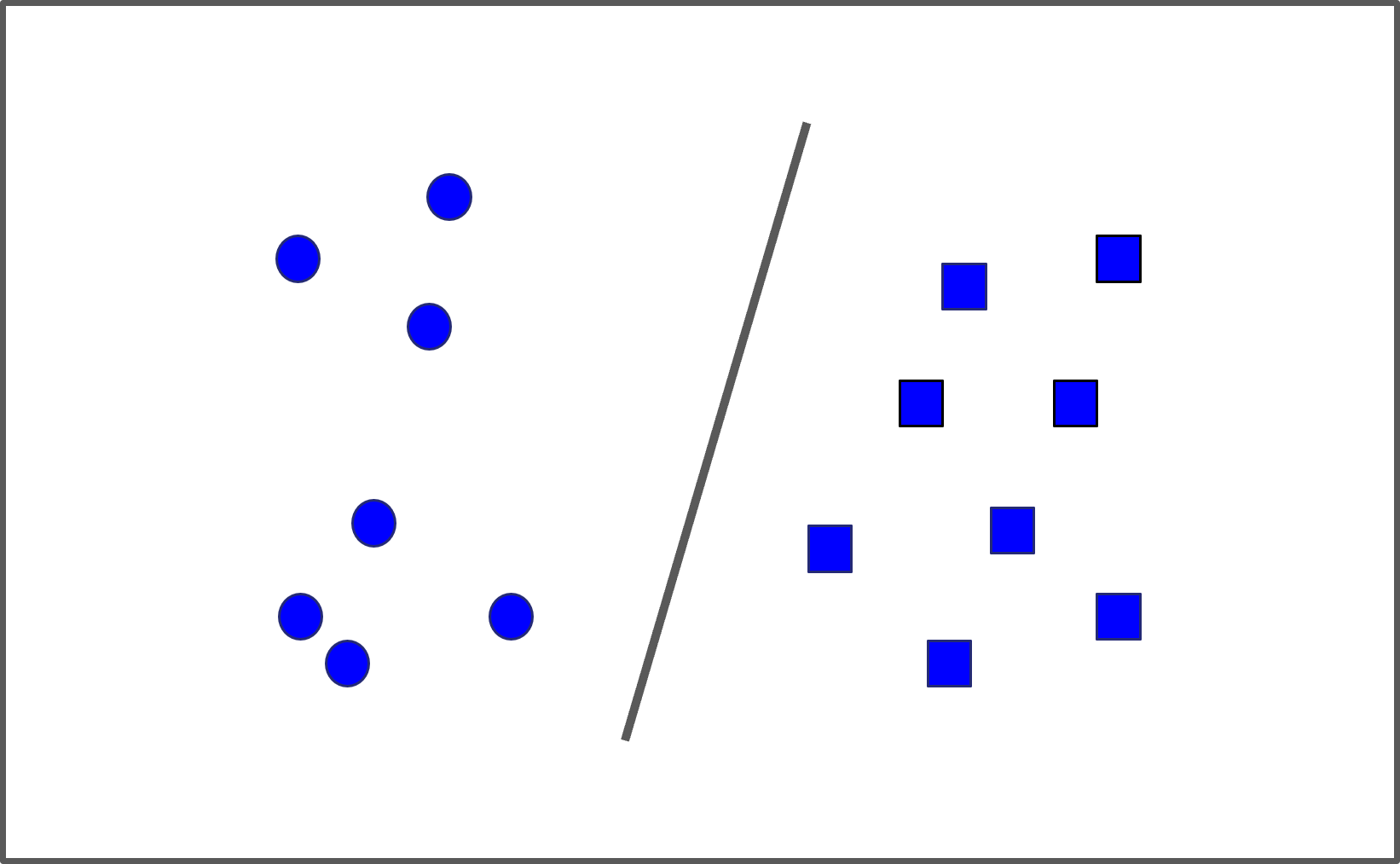}
    \caption{Test (Spurious)}
    \label{fig:spurious_test}
    \end{subfigure}
    \begin{subfigure}[b]{0.24\linewidth}
        \centering
        \includegraphics[width=\linewidth]{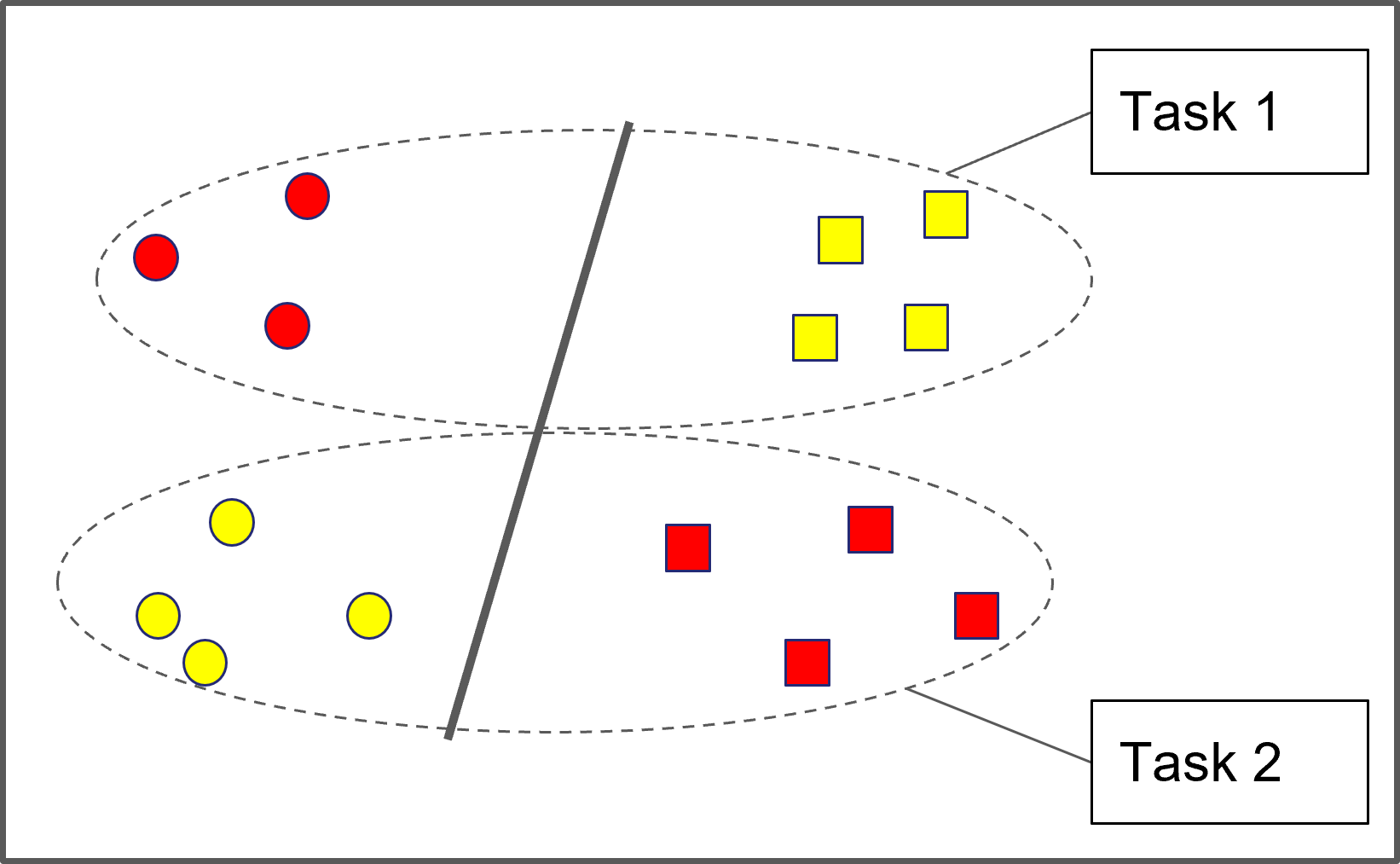}
    \caption{Train (local spurious)}
    \label{fig:virtual_train}
    \end{subfigure}
    \begin{subfigure}[b]{0.24\linewidth}
        \centering
        \includegraphics[width=\linewidth]{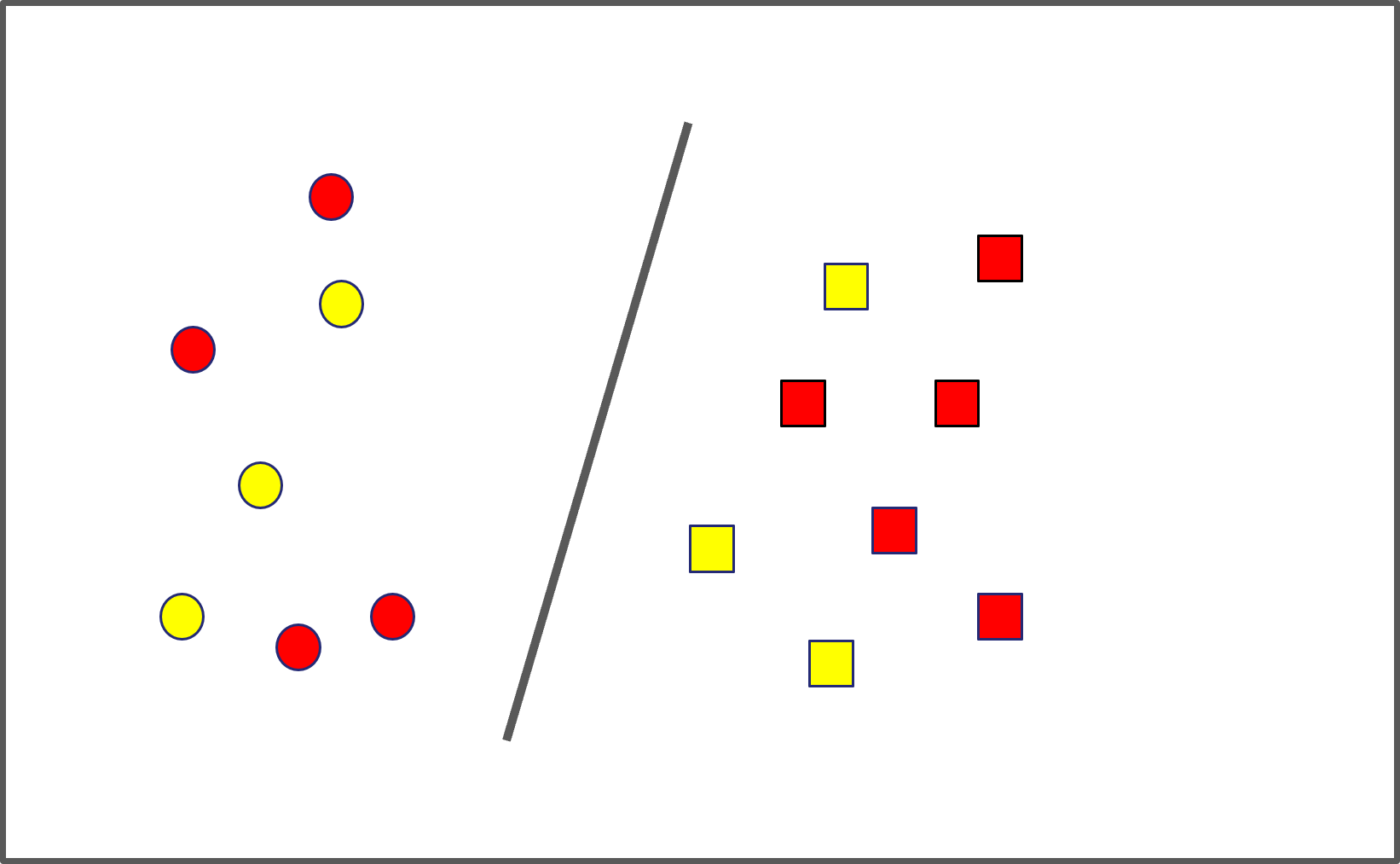}
    \caption{Test (local spurious)}
    \label{fig:virtual_test}
    \end{subfigure}
    
    \caption{\checked{\textbf{Spurious features and local spurious features.} If the task is to distinguish the squares from the circles. In Fig.~\ref{fig:spurious_train} and \ref{fig:spurious_test}, the color is a spurious feature because there is a covariate shift between train and test data. In Fig.  \ref{fig:virtual_train} and \ref{fig:virtual_test}, we observe two tasks of a domain-incremental scenario, the colors are locally spurious in tasks 1 and 2. Even if there is no significant covariate shift between train and test full data distribution, colors appear discriminative while looking at data within a task.}}
    \label{fig:illustration}
\end{figure*}

This paper investigates both the problem of spurious features (with covariate shift) and local spurious features (without covariate shift) in CL as shown in Fig.~\ref{fig:illustration}. 
Our contributions are: 
(1) We propose a methodology to highlight the problems of spurious features and local spurious features in continual learning.
(2) We create a binary CIFAR10 scenario \textit{SpuriousCIFAR2} inspired by colored MNIST to experiment with spurious correlations.
(3) We propose a modified version of Out-of-Distribution (OOD) generalization methods for continual learning and evaluate them on \textit{SpuriousCIFAR2}.
(4) We identify local spurious features as a core challenge for continual learning algorithms along with catastrophic forgetting.

\checked{We expect this paper to be a significant step toward a better understanding of the continual learning problem. It also proposes baselines and settings for further investigation of spurious features in continual learning.}

\section{Related Work}
\label{sec:related}

\checked{In large part of the continual learning bibliography, algorithms assume that to avoid catastrophic forgetting, they should not increase the loss on past tasks \cite{kirkpatrick2017overcoming,Ritter18Online}. It leads to the definition of interference/forgetting of \cite{riemer2018learning}: $\frac{\partial L(x_i,y_i)}{\partial \theta} \cdot{} \frac{\partial L(x_j,y_j)}{\partial \theta} < 0$, 

$\forall (x_i,y_i) \in \mathcal{T}_i$ and $\forall (x_j,y_j) \in \mathcal{T}_j$ with $j>i$, $<\cdot{}>$ is the dot product operator.
Following this definition, increasing the loss on past tasks necessarily leads to a performance decrease.}
%
However, in the presence of spurious features, the algorithm might have learned spurious/local features that need to be forgotten to improve performance as discussed in appendix \ref{ap:sec:discussion}. Hence, the loss needs to be temporarily increased to reach a more general solution, and optimizing the interference equation could be counterproductive.
On the same line, the presence of spurious features is adversarial to most continual regularization strategies. Indeed, if we measure weight or example importance with Fisher information, high importance will be given to weights using spurious features, and regularization will penalize their modification. 
Regularization will force the memorization of features that generalize poorly to the test set.

Vanilla rehearsal or generative replay can be a good solution to avoid forgetting meaningful information and dealing with spurious features. Indeed, by replaying old data, algorithms simulate an independent and identical distribution (iid) and avoid local spurious feature problems. Replay methods have been shown in the bibliography to be efficient and versatile even in their most straightforward form \cite{prabhu2020gdumb}. Notably, all current state-of-the-art approaches on ImageNet use replay \cite{douillard2020podnet,zhao2020maintaining}.

\checked{The research field that usually deals with spurious correlations is the out-of-distribution (OOD) generalization field.
This field has received a lot of attention in recent years, especially since the Invariant Risk Minimization (IRM) \cite{arjovsky2019invariant} paper.
OOD approaches target training scenarios where there are several training environments within which different spurious features correlate with labels. The goal then is to learn invariant features among all environments to build an invariant predictor in all training environments and potentially any other \cite{arjovsky2019invariant,ahuja2021invariance,sagawa2019distributionally,pezeshki2020gradient}. This paper will adapt some of those approaches for continual learning to evaluate how those approaches can deal with sequences of tasks.}

\section{Problem Formulation}
\label{sec:problems}

\checked{This section introduces the spurious features problems in a sequence of tasks. The goal is to present the key concepts and challenges for continual feature selection. 
First, we will describe general, local, and spurious features. Secondly, we will present the different cases of spurious features.}

\begin{wraptable}{R}{0.5\textwidth}
\caption{\checked{Summary of characteristics of the types of features. For a feature $z$ of a class $c$, we denote if it verify (\ref{eq:discr}) on different data setting, a single task $\mathcal{T}_t$, the whole scenario $\mathcal{C}_{T}$, the test set $\mathcal{D}_{te}$.}}
\label{tab:variables}
\centering
\resizebox{\linewidth}{!}{
\begin{tabular}{@{}lp{1cm}p{1cm}p{1cm}@{}}
\toprule
\textbf{Name} & $\mathcal{T}_{t}$ & $\mathcal{C}_{T}$ & $\mathcal{D}_{te}$ \\ \midrule
Good Feature ($z_+$) & \checkmark & \checkmark & \checkmark\\
Spurious Feature ($z_{spur}$) & \checkmark & \checkmark & $\times$\\
Local Feature ($z_{loc}$) & \checkmark & ? & ?\\
Local Spurious Feature ($z_{spur:t}$) & \checkmark & $\times$ & $\times$ \\ \bottomrule
\end{tabular}}
\end{wraptable}

\subsection{General, Local, Spurious and Local Spurious Features}
\label{sub:features}

\textbf{General Formalism: }
\checked{We consider a continual scenario of classification tasks.
We study a function $f_{\theta}(\cdot)$, implemented as a neural network, parameterized by a vector of parameters $\theta \in \mathcal{R}^{p}$ (where p is is the number of parameters) representing the set of weight matrices and bias vectors of a deep network. In continual learning, the goal is to find a solution $\theta^*$ by minimizing a loss $L$ on a stream of data formalized as a sequence of tasks $[\mathcal{T}_0, \mathcal{T}_1, ..., \mathcal{T}_{T-1}]$, such that $\forall (x_t,y_t)\sim \mathcal{T}_t$ ($t \in [0, T-1]$), $f_{\theta^*}(x)=y$. We do not use the task index for inferences (i.e. single head setting).}

\checked{To describe the different types of features, let $z$ be a feature and $x\sim \mathcal{D}$ a datum point in dataset $\mathcal{D}$. We define $w(.)$ a function which returns 1 if $z$ is in $x$ and 0 if not. $w(.)$'s output is binary for simplicity.
Then, for all data with a label $y$ in the dataset $\mathcal{D}$, we can compute the correlation $c(\mathcal{D}, z, y) = correlation(w(z,x)=1, Y=y)$, which estimates how a feature correlates with the data of a given class}.
We can then define discriminative features as:

$z$ is discriminative for class $y$ in $\mathcal{D}$ if:
\begin{equation}
    \forall y' \in \mathcal{Y}, y \neq y' \quad c(\mathcal{D}, z, y) \gg c(\mathcal{D}, z, y')
\label{eq:discr}
\end{equation}
\checked{$\mathcal{Y}$ is the set of classes in $\mathcal{D}$.
In other words, $z$ is discriminative for $y$ if it correlates significantly more to $y$'s data than to the data of any other class.
Then a good feature $z_{+}$ for a class $y$ respects (\ref{eq:discr}) for training data $\mathcal{D}_{tr}$ and test data $\mathcal{D}_{te}$.}
%

\textbf{Spurious Features and Local Spurious Features: }

\checked{A spurious feature $z_{spur}$ for a class $y$ respects (\ref{eq:discr}) for training data $\mathcal{D}_{tr}$ but not for test data $\mathcal{D}_{te}$.
A spurious feature is well correlated with labels in training data but not with testing data.}
\checked{Hence, learning from $z_{spur}$ may offer a low training error but high test error. The presence of $z_{spur}$ is due to a covariate shift between train and test distribution which changes the feature distribution.}

In continual learning, the covariate shift between train and test $z_{spur}$ may also lead to poor generalization. Further, the features can be locally spurious, e.g., they correlate well with labels within a task but not within the whole scenario. We name them \textit{local spurious features}. We illustrate the difference between spurious features and local spurious features in Figure \ref{fig:illustration}.
 
At task $t$, A local spurious feature $z_{spur;t}$ respects (\ref{eq:discr}) for a class $y_t$ in task $\mathcal{T}_t$, but not for the whole scenario $\mathcal{C}_T$. 

$z$ is a local spurious feature for a class $y$ in $\mathcal{T}_t \sim  \mathcal{C}_T$, with $t \in \llbracket 0 , T-1 \rrbracket$:
\begin{equation} 
\begin{split}
    \mbox{if } \forall~y' \in \mathcal{Y}_t, y \neq y' \quad c(\mathcal{T}_t, z, y) \gg c(\mathcal{T}_t, z, y')\\ 
    \mbox{but } \exists~y'' \in \mathcal{Y}, y \neq y'' \quad c(\mathcal{C}_T, z, y) \not \gg c(\mathcal{C}_T, z, y'')
\label{eq:discr}
\end{split}
\end{equation}
$\mathcal{Y}_t$ is the set of classes in task $\mathcal{T}_t$ and $\mathcal{Y}$ is the set of classes in the full scenario $\mathcal{C}_T$ composed of $T$ tasks.

In other words, a local spurious feature $z_{spur;t}$ correlates well with a label on the current task but not on the whole scenario. $z_{spur;t}$ can be extended from a single task $\mathcal{T}_t$ to all task seen so far $\mathcal{T}_{0:t}$ without loss of generality.

\textbf{Global vs Local Solution: }
We assume that machine learning models solve tasks by learning to detect (or select) features that correlate well with labels.
Then, while learning on a task $t$, we can distinguish a local solution $\theta_t^*$, satisfying for the current task $\mathcal{T}_t$, from a global solution $\theta_{0:T}^*$ that is satisfying for whole scenario $\mathcal{C}_T$ (past, current, and future tasks).

Similarly, we can differentiate local and global features, contributing to local and global solutions. The global features are the good features $z_{+}$ that are useful for the solution of the scenario. Unfortunately, at time $t$, we can not know if a feature is part of $z_{+}$ without access to the future. Therefore, algorithms should learn with their current data but update their knowledge afterwards, given new data. 
%
%
For example, in classification, the discriminative features for a given class depend on all the classes. Therefore, when new classes arrive, discriminative features can become outdated in class-incremental scenarios.

\subsection{Spurious Correlations: Different cases}
\label{sub:cases}

\checked{We defined the different types of features in the previous section. We can now identify different cases among the spurious correlation between features and labels.}

\textbf{Data Observability:}\\
    \quad $\bullet$ \textbf{Fully observable data} \cite{javed2020learning} \checked{The global features' are always present and always observable. In this case, we can assume that features in data that do not last are spurious, and we can learn to ignore them.}\\
    \quad $\bullet$ \textbf{Partially observable data} \checked{The global features' presence is not invariant. In this case, features that do not last can either be spurious or good}.

\textbf{Noisiness of Spurious Features:}\\
    \quad $\bullet$ \textbf{SFs are irrelevant for other classes} they can be ignored completely without affecting the learning process. We will refer to them as noisy spurious features.\\
    \quad $\bullet$ \textbf{SFs are good features for other classes},
    \checked{e.g., for classification, the color can be a spurious feature for some classes and valuable for others. We can not ignore those features since it could lead to poor performance in other classes.}

\checked{In our experiment, we propose settings with fully observable and partially observable data. Our spurious features are noise in our domain incremental experiments and true features in our class-incremental experiments. Nevertheless, we do not exploit information about spurious feature types to design approaches. }

\section{Spurious Features}

This experimental section studies how continual learning algorithms can deal with spurious features. We design a scenario with spurious features that change at each task. We create a set of scenarios with gradual correlations between spurious features and labels. We evaluate various baselines to assess continual learning capabilities (Sec. \ref{subsub:correlation}) in such scenarios. We also experiments with potential solutions to deal with spurious correlation. 

\subsection{Setting}

\begin{wrapfigure}{r}{0.4\textwidth}
\begin{tikzpicture}[main_node/.style={circle,draw,minimum size=5em,inner sep=10pt]}]

    \node[main_node] (1) at (0,0) {$t$};
    \node[main_node] (2) at (0, -2)  {$Y$};
    \node[main_node] (3) at (2, 0) {$z_{spu}$};
    \node[main_node] (4) at (2, -2) {$z_{+}$};
    \node[main_node] (5) at (4, -1) {$X$};

    \draw [-stealth] (1) -- (3);
    \draw [-stealth] (2) -- (4);
    \draw [-stealth] (3) -- (5);
    \draw [-stealth] (4) -- (5);
\end{tikzpicture}
    \caption{ Data generation model of SpuriousCIFAR2 scenario. The good features $z_{+}$ are generated from the labels and $z_{spu}$ are generated from the task label.}
\label{tikz:causal_graph}
\end{wrapfigure}
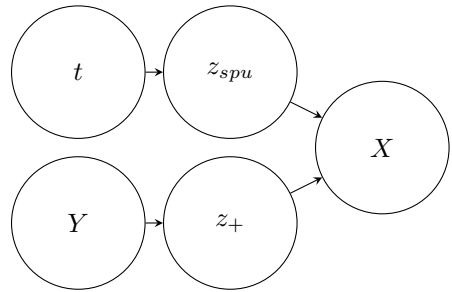


\checked{The proposed benchmark is similar to colored MNIST \cite{arjovsky2019invariant} with CIFAR10. We convert the ten-way classification dataset into a binary dataset. The new classes are \say{transportation means versus not transportation means}, i.e., cars, trucks, ships, airplanes, and horses versus the other classes: birds, cats, dogs, deers, frogs. The goal is to have a simple setting more challenging than colored MNIST with features that are built upon all three color channels.}

\checked{The spurious feature is a square of color. We sample two colors randomly (one per class) and add a $2\times2$ pixels square randomly positioned in the images (example Fig.~\ref{ap:sec:samples}). We can vary the percentage of images with the spurious feature to reduce or grow the correlation between the spurious feature and the labels. A correlation of $1$ means that all images have a colored square.}

\checked{The scenario is then a sequence of SpuriousCIFAR2 datasets with different colors of spurious features. The test set of each scenario is the binarized version of CIFAR10 test set without any spurious features. We created the sequence of tasks using the \textit{Continuum} library \cite{douillard2021continuum}.}
We illustrate two environments and the test set in Fig.~\ref{ap:sec:samples}, and the data generation process in Fig.~\ref{tikz:causal_graph}.
The good features are invariant through tasks if each task contains the full dataset (cf experiments \ref{subsub:correlation}).

\checked{\textbf{Setting Goal: } This setting is made to highlight how spurious features can disrupt continual learning algorithms and discuss the problems that spurious correlation can bring to existing approaches. Moreover, it evaluates the capacity of algorithms to question and modify their past knowledge to improve test accuracy. }

\subsection{Approaches}

\checked{First, we experiment with a classical vanilla replay method and simple finetuning. 
For replay, the replay buffer is constructed by randomly selecting $N$ samples per class. The buffer is then sampled to keep class distribution balanced over all tasks. Balancing samples distribution over classes is made to avoid the challenge of training on imbalanced datasets (cf Sec. \ref{ap:sec:algo_sampling} in the appendix for details).}

\checked{Among the existing OOD approaches, we compare continual versions of IRM \cite{arjovsky2019invariant} and the state-of-the-art OOD classification methods
IB-ERM, IB-IRM \cite{ahuja2021invariance}, GroupDRO \cite{sagawa2019distributionally} and Spectral Decoupling \cite{pezeshki2020gradient}.
OOD approaches are algorithms designed to be trained on multiple environments in a multi-task fashion. They propose different regularization strategies designed to ignore spurious features and learn invariant features. The invariant features are present in all environments and are assumed to be good features. The goal of the OOD approaches is to learn an invariant predictor that would ignore spurious features and rely only on invariant features.
The continual version of OOD approaches simulates multiple environments by replaying data of past tasks. The adaptation of all those methods is then to add a replay buffer to the algorithms to train continually the baseline through the sequence of tasks. The replay buffer simulates the growing number of environments for the OOD approaches.
 We choose empirically a replay buffer storing 100 samples per class for both OOD approaches and vanilla replay (cf appendix \ref{ap:sec:HPs} for hyper-parameters selection protocol).}

\subsection{Experiments}

\subsubsection{Problem Highlights}
\label{sub:highlights}
\begin{wrapfigure}{r}{0.25\textwidth}
\begin{minipage}{0.25\textwidth}
    \centering
    \includegraphics[width=\linewidth,trim={0 1cm 0 0},clip]{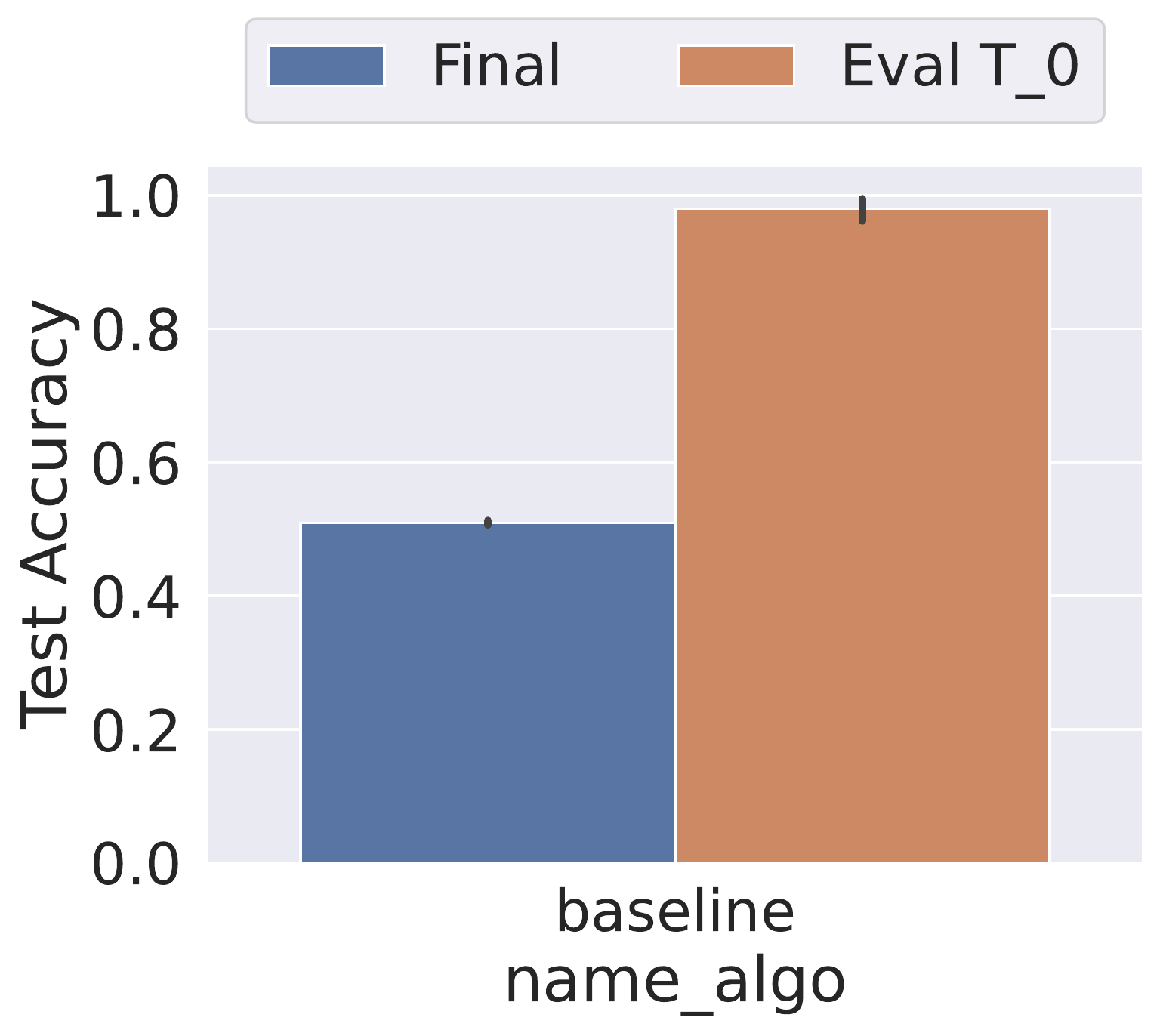}
    \caption{Overfitting the spurious features on a single task setting}
    \label{fig:overfitting}
\end{minipage}

\begin{minipage}{0.25\textwidth}
    \centering
    \includegraphics[width=\linewidth]{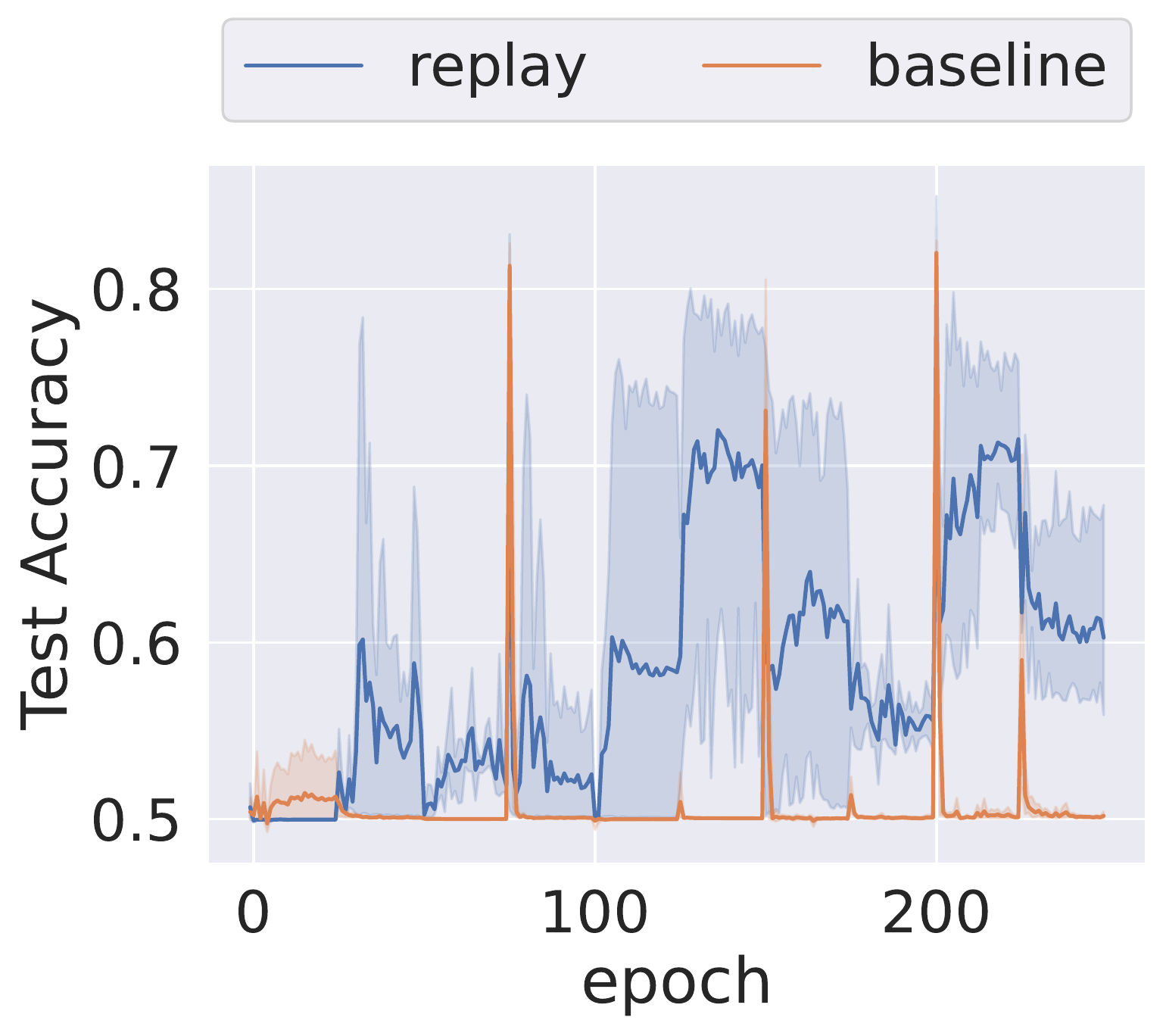}
    \caption{Generalization and forgetting in a sequence of tasks}
    \label{fig:instability}
\end{minipage}
\end{wrapfigure}
\textbf{Overfitting the spurious features:}
\checked{To assess that algorithm overfit on the spurious features, we compare the test accuracy without spurious features (final test set) with the test accuracy with spurious features (evaluation set).
If the test accuracy is good with spurious features and bad without, the algorithm overfits the spurious features.
Fig.~\ref{fig:overfitting}, show exactly this phenomenon, test accuracy on task 0 is near-perfect accuracy, while on the final test set, the accuracy is near-random prediction.
This figure shows that the artificial spurious features cause the expected learning behaviors: the model overfits them spontaneously and generalizes poorly.}

\textbf{Instability: }
\checked{In Fig.~\ref{fig:instability}, we assess the test accuracy at each epoch over the whole sequence of 10 tasks. This figure indicates two interesting pieces of information. First, even in the 100\% spurious correlation, i.e. all images have a square of colors, baseline models can learn at some point a good solution. Secondly, even when they learn a good solution, they are very unstable and can easily forget a good solution.
To lighten the influence of instability on the evaluation metrics, we report the average test accuracy after each task, which we note $\Omega$ instead of reporting the final test accuracy.}

\textbf{Comparing CL Baselines with OOD Baselines: }
\begin{wrapfigure}{R}{0.3\textwidth}
    \centering
    \includegraphics[width=\linewidth,trim={0 1cm 0 0},clip]{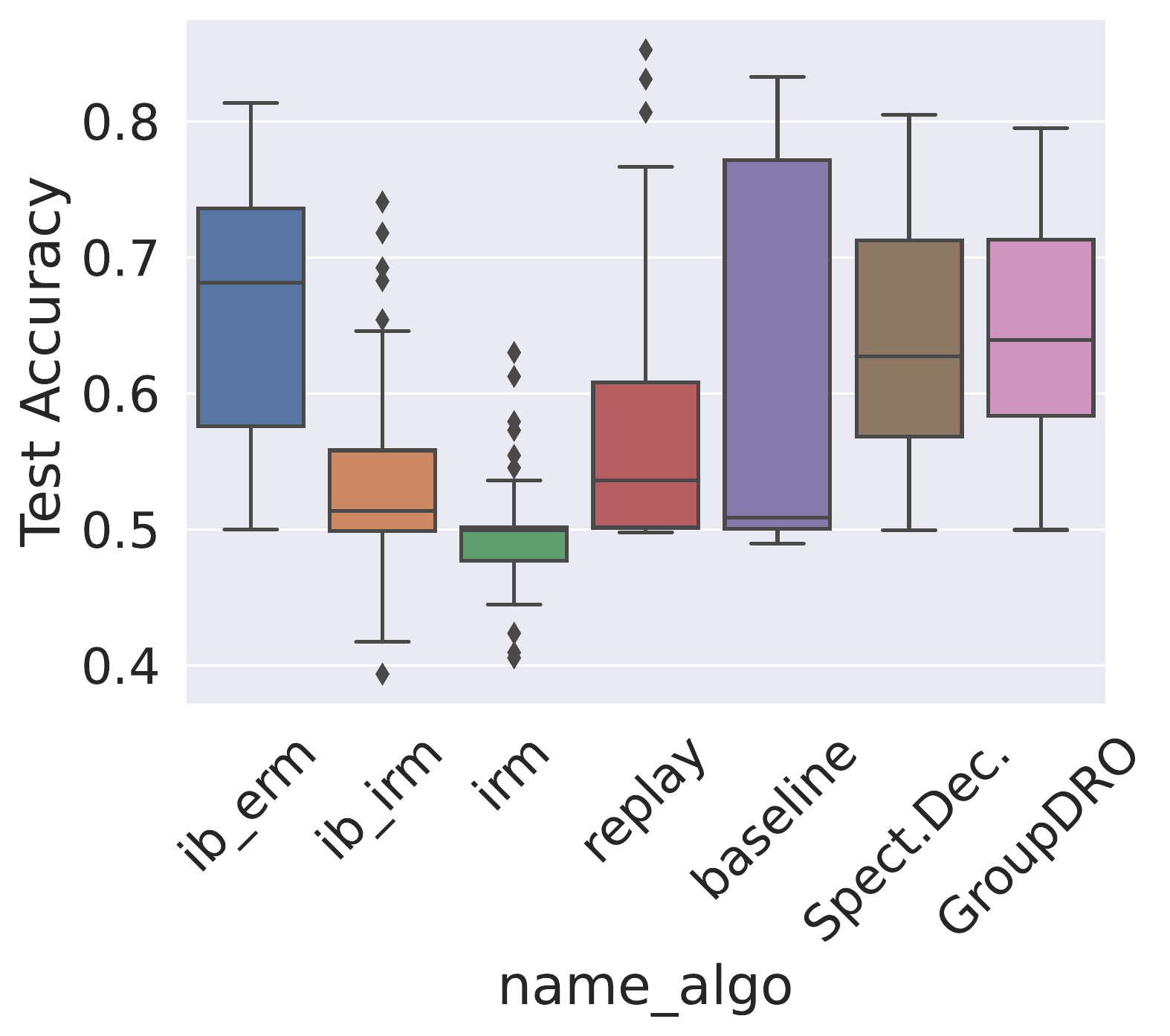}
    \caption{Accuracy with local 100\% correlation between features and labels}
    \label{fig:initial_spurious}
\end{wrapfigure}
\checked{We now assess how baselines adapted from the OOD field behave in the 100\% spurious correlation setting.
 In our experiments, \textit{IB-ERM}, \textit{SpectralDecoupling} and \textit{GroupDRo} show some interesting improvement over rehearsal and finetuning baselines ( Fig.~\ref{fig:initial_spurious}). On the other hand, \textit{IRM} and \textit{IB-IRM} performed poorly.}

\checked{Those experiments show us that we can have some improvement over replay and finetuning baselines when there is $100\%$ correlation between spurious features and labels. Note that there is much variance in results because of the instability mentioned earlier. Moreover, the improvements stay far from a satisfying accuracy. Indeed, the average performance of the baselines on CIFAR2Spurious is below $70\%$ of accuracy. Meanwhile, we trained a model on the CIFAR2 dataset without spurious features and reached $96.73\%$ of accuracy.
The next experiments will analyze performance while gradually growing the spurious correlations. }

\subsubsection{Influence of Spurious Correlation}
\label{subsub:correlation}

\checked{In this section, we aim to answer the question \say{how does the level of spurious correlation influence learning algorithms?}.
From a stream of tasks, we can expect that the 100\% spurious correlation is quite rare. Hence, we will investigate the setting of lower spurious correlation in this set of experiments. We study the $25\%$, $50\%$ and $75\%$ spurious correlation cases along the $100\%$ correlation.}

\begin{figure}
    \centering
    \includegraphics[width=0.7\linewidth]{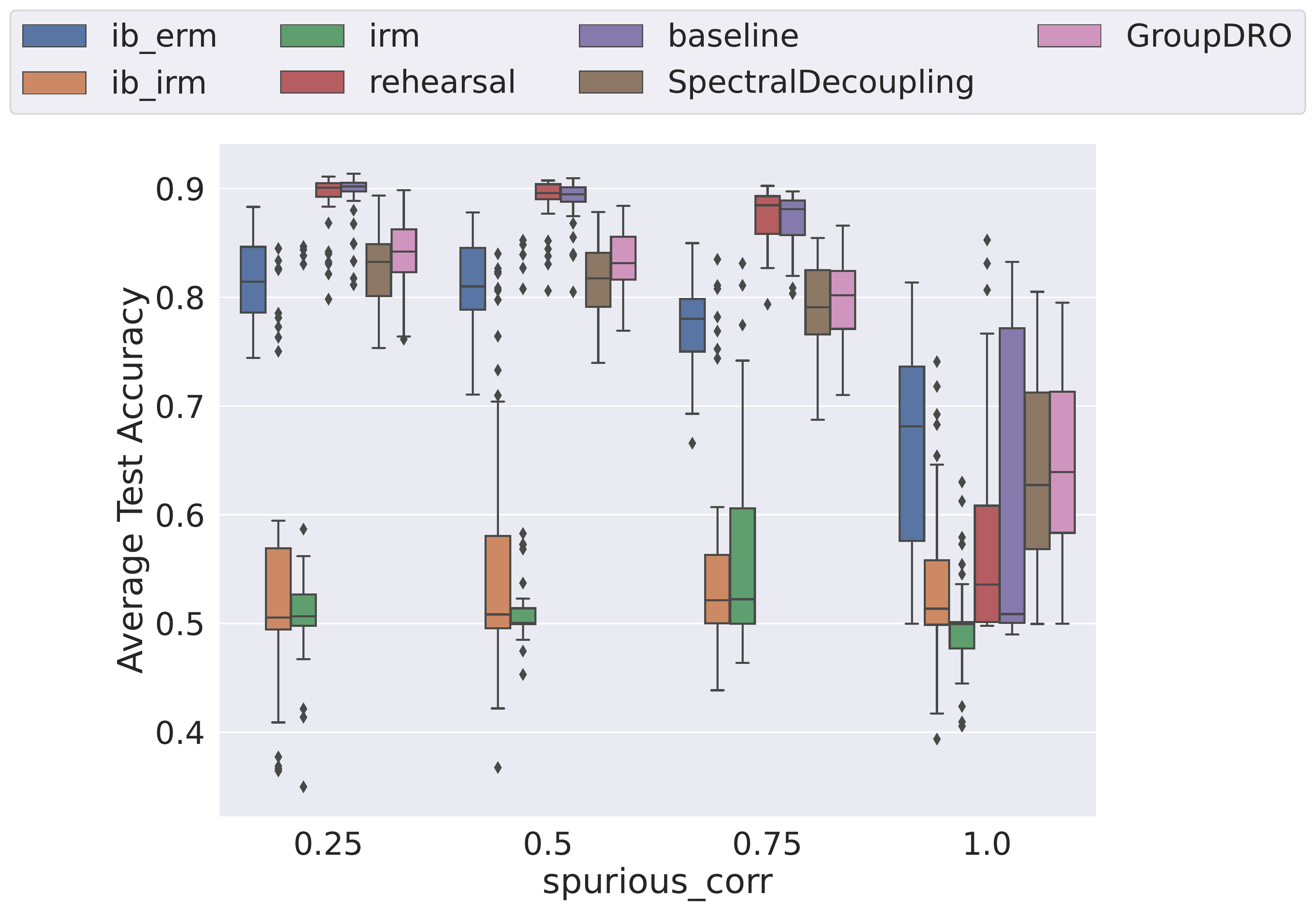}
    \caption{Averaged accuracy $\Omega$ on 10 tasks over various amount of spurious correlation between spurious features and labels.}
    \label{fig:averaged_spurious}
\end{figure}
Fig.~\ref{fig:averaged_spurious} show that lowering the spurious correlation makes rehearsal and finetuning (baseline) the best approaches.
Those results indicate us the OOD baselines are most interesting in the very high spurious correlation setting but are not very interesting when the spurious correlation is lower or equal to $75\%$.
 It might seem counter-intuitive that finetuning is over the best baseline in a continual learning setting. Still, only the spurious correlation change in our scenario, so the global features needed to solve all tasks are in each task. It is then possible that with a lower spurious correlation finetuning works well.
 %


The scenario could be made harder by only keeping a subset of the dataset in each task, i.e., lowering the support of the data distribution (cf in appendix \ref{ap:sec:support}). We can note that lowering the support does not make only the task harder because there are fewer data per task but also because it removes the full observability of data.
It means that good/global features are no longer always observable and that good features can be available in some tasks but not others. Nevertheless, our results on the support showed that it has a low influence on the results. This is probably because replay makes everything observable at the last tasks and overcomes partial support challenges. It is similar to creating several environment with separate split of the full distribution as in DomainBed datasets.

\subsubsection{Potential Solutions to Lower Impact of Spurious Features}
\label{sub:solution}

\begin{wrapfigure}{R}{0.5\textwidth}
    \centering
    
    \begin{subfigure}[b]{0.45\linewidth}
        \centering
        \includegraphics[width=\linewidth,trim={0 1cm 0 0},clip]{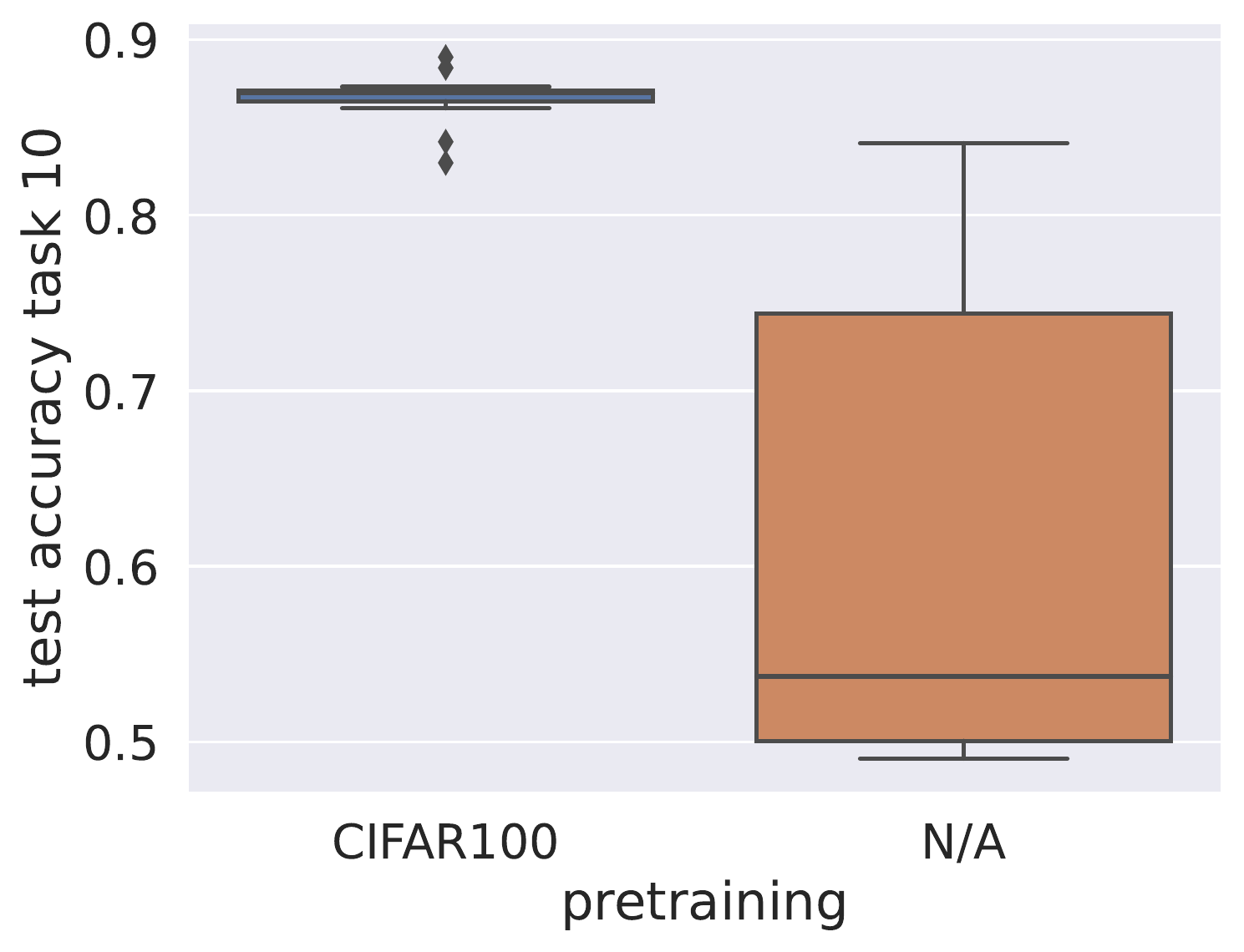}
        \caption{Pre-training}
        \label{fig:pre-training}
    \end{subfigure}
    \begin{subfigure}[b]{0.45\linewidth}
        \centering
        \includegraphics[width=\linewidth,trim={0 1cm 0 0},clip]{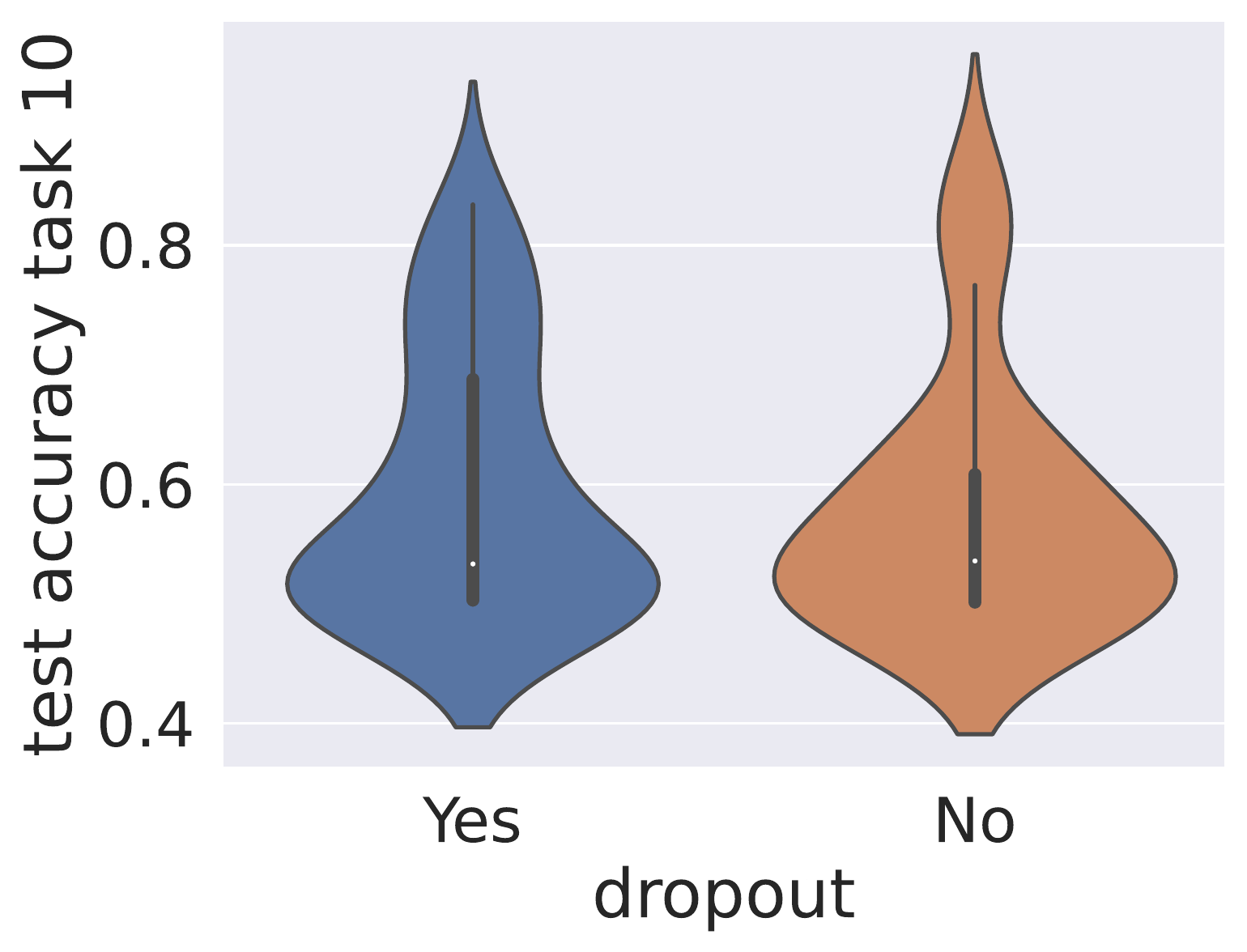}
        \caption{Dropout}
        \label{fig:dropout}
    \end{subfigure}
    
    \caption{Canceling noisy spurious features with pre-trained models.
    }
\end{wrapfigure}

\checked{A solution to prevent models from relying too much on spurious features is (1) to force them to learn/use more features or (2) to try to ignore the spurious features.
We experiment with (1) by using a regularization strategy to maximize the number of features selected. This regularization is different from regularization methods designed to not forget. We experiment with (2) by using a model pre-trained on a trusted task. The idea of the pre-trained model is that it will ignore noisy features, such as our spurious features.}

\textbf{Using a pre-trained model: }

\checked{We can use a pre-trained model on a trusted data source to ignore spurious features.
For example, the spurious feature in our setting is noise, so if we use a pre-trained model on a known dataset such as CIFAR100, we can significantly improve results.
This approach, experimented in Fig.~\ref{fig:pre-training}, shows clearly that using a pre-trained model can erase the problem of noisy spurious correlation.
This solution is convenient, but it assumes we have a compatible trusted set of data (or a trusted model) and that the spurious features are noisy.}

\textbf{Maximizing the amount of features selected: }

If we can not use the previous solution, another potential solution is to learn to select as many features as possible that could help solve the problem. A famous solution to maximize the features learned is \textbf{dropout}. Dropout randomly replaces some activations value with a zero for inference to force the model to learn more robust features. It has notably been widely experimented in continual learning \cite{Goodfellow13,mirzadeh2020dropout}. 
We experimented $0.25$, $0.5$, and $0.75$ amount of dropout just before the last linear layer. However, unfortunately in our experiments, it did not show any improvement with dropout than without (cf. Fig.~\ref{fig:dropout}).

\checked{Nevertheless, on a similar idea as dropout, the \textbf{spectral decoupling} approach is designed to address the gradient starvation problem. The gradient starvation problem arises when the loss is minimized by capturing only a subset of features relevant to the task, despite the presence of other predictive features that fail to be discovered \cite{pezeshki2020gradient}. Spectral decoupling is designed to discover supplementary features even with minimal train error. As dropout, it enables the possibility to learn additional features that could help to improve the test error. The experiment in section \ref{sub:highlights} illustrated in Fig.~\ref{fig:initial_spurious}  indeed shows that in the 100\% spurious correlation experiment, this strategy greatly improves simple rehearsal proving the potential of the idea.}

\checked{We proposed trivial solutions to illustrate how supplementary knowledge or assumptions on the spurious features might help prevent or fix bad learning behavior. However, it would probably not work as easily in a setting with more complex spurious features.}

\checked{Experiments in sections \ref{subsub:correlation} and \ref{sub:solution} investigate how algorithms learning continually can deal with spurious features. We created a benchmark with spurious features and empirically investigated algorithms' performance when varying the correlation of spurious features with labels. In the next section, we will investigate local spurious features, which are a type of spurious feature specific to continual learning. We will investigate if those features may cause a performance decrease in continual learning algorithms.}

\section{Local Spurious Features}
\label{sec:local}

\checked{In this section, we want to show that local spurious features lead to a performance decrease.
Therefore, we design a setting where algorithms learn local solutions from independent tasks, and then we test them with all tasks together.
It makes it possible to estimate if the features learned within a task are generalizable to the full scenario or if they were only locally correlated with labels. The setting is designed such that past weights can not be modified, \say{forgetting} can not cause performance decrease and interfere with the local spurious features problem.}

 \subsection{Local Spurious Features Setting}
 \label{sub:local_setting}

\checked{We experiment with original datasets without modification to create class-incremental settings as in most continual learning literature. 
We use CIFAR10, OxfordPet, OxfordFlowers and CUB200 datasets with pretrained models: a resnet model on CIFAR100 \footnote{\textit{https://github.com/chenyaofo/pytorch-cifar-models} (BSD 3-Clause License)} for CIFAR10 and for the other datasets we use VGG, Resnet, Alexnet and googlenet pretrained on Imagenet from torchvision library \footnote{\textit{https://pytorch.org/vision/stable/models.html} (BSD 3-Clause License)}.
We create the scenarios by splitting each dataset into 5 tasks with disjoint sets of classes.}

\checked{We assume that the pre-trained models provide a feature space sufficient for the continual downstream scenarios. They are used frozen with a linear classifier on top. We also assume that algorithms select features that correlate well with labels to solve the current task.
The goal is then to show that while learning, classifiers select features that correlate well with labels locally within a task but not in the full scenario. If it is the case, this will lead to a good performance on the tasks but not in the full scenario. The training is a simple finetuning without any replay or regularization. }

We use a multi-head approach to see if the classifier learns local features. Hence, while training, we only apply the softmax function to the outputs of the current task's classes. We note the test performance in \textit{multi-head} $A_{te-local\_softmax}$. This first evaluation measures the mean performance on each task separately. It estimates the capabilities of features selected to solve tasks independently.
After each task, we freeze weights of past heads to avoid forgetting.
After training, we compare the test performance with the same classifier (i.e., the classifier trained with multi-head) but with the softmax applied on all outputs $A_{te-global\_softmax}$, similarly to \textit{single head} inference. 
This second evaluation estimates the generalization capabilities of the features selected.
Hence, if classifiers select features that are not generalizable, $A_{te-local\_softmax}$ should be significantly bigger than  $A_{te-global\_softmax}$.

Nevertheless, apart from the problem of feature selection, a gap between $A_{te-local\_softmax}$ and $A_{te-global\_softmax}$ could be explained by  (1) the difference in difficulty of both evaluation (multi-head and single head), (2) an unbalance of bias and norms from different heads that could lead to bad performance in \textit{single head} \cite{lesort2021continual} (more details about this problem in appendix \ref{ap:sec:bias_norm}). However, if neither (1) or (2) are sufficient to explain the performance gap, then the feature selection was not generalizable, i.e., the multi-head training selected spurious local features. 

\checked{\textbf{Performance gap: } the comparison between single head and multi-head is biased because the first is one 10-way classification (harder) while the latter is the addition of five binary classifications (simpler). 
To estimate the difference of difficulty between both, we added a non-parametric method, \textit{MeanLayer} as in \cite{lesort2021continual} which is an nearest mean classifier (NMC). 
There is no feature selection in MeanLayer. The classifier only uses the mean of the features of each class. There is no feature selection then no problem with local features selection. The difference in performance with MeanLayer in multi-head and the single head is then a good proxy to estimate the difference in the difficulty of both evaluations.}

\checked{\textbf{Unbalance of bias and norms:} we compare the linear layer performance with the weightnorm layer from \cite{lesort2021continual}. This layer does not use norm and bias for inference and is then insensitive to such imbalance (details in appendix \ref{ap:sec:bias_norm}).}

\checked{We note that forgetting is by design impossible here since the features extractor and the other heads are frozen. Hence, forgetting can not explain drops in performance.}

\begin{figure}[h]
    \centering
    \includegraphics[width=0.24\linewidth,trim={0 1cm 0 0},clip]{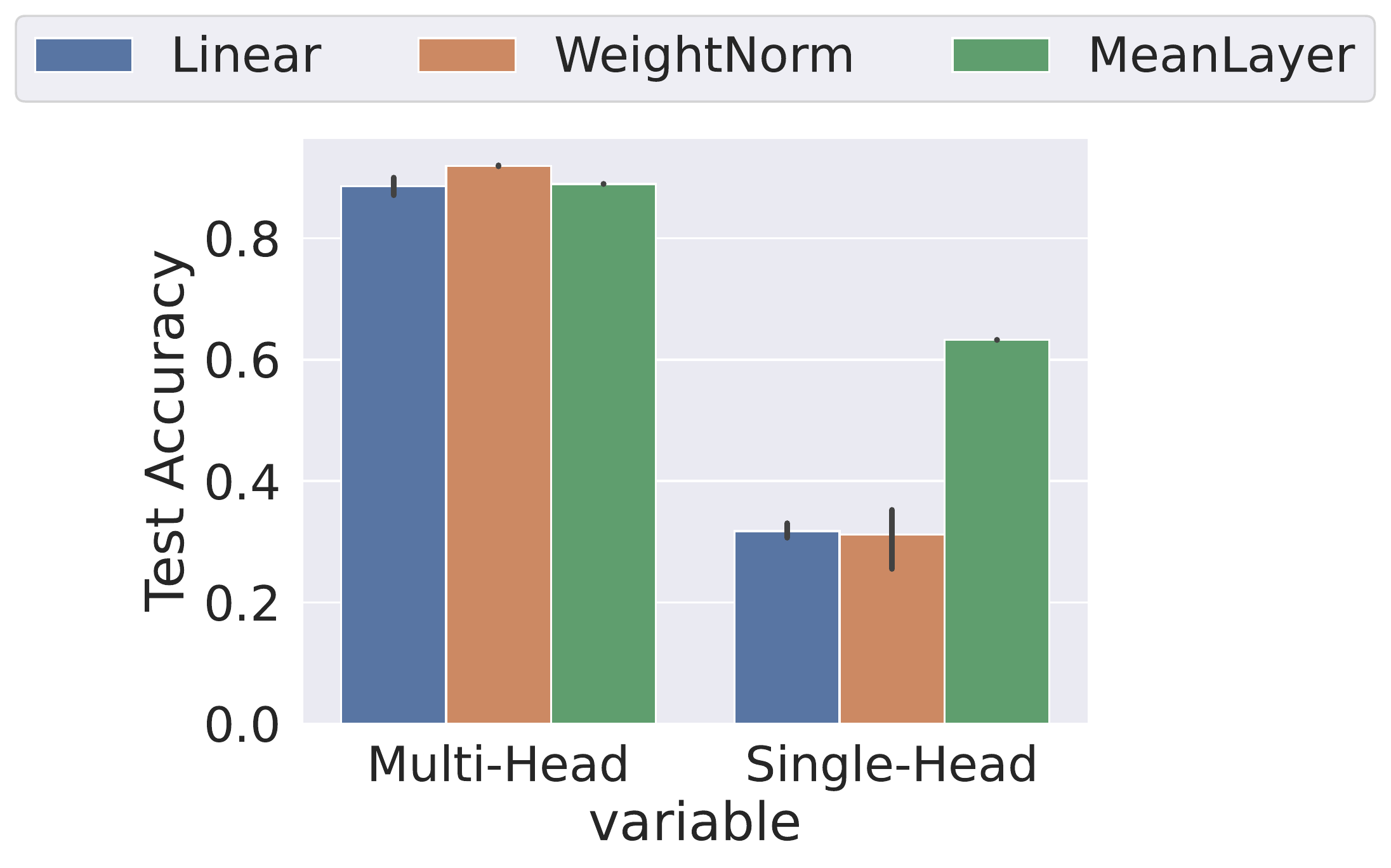}
    \includegraphics[width=0.24\linewidth,trim={0 1cm 0 0},clip]{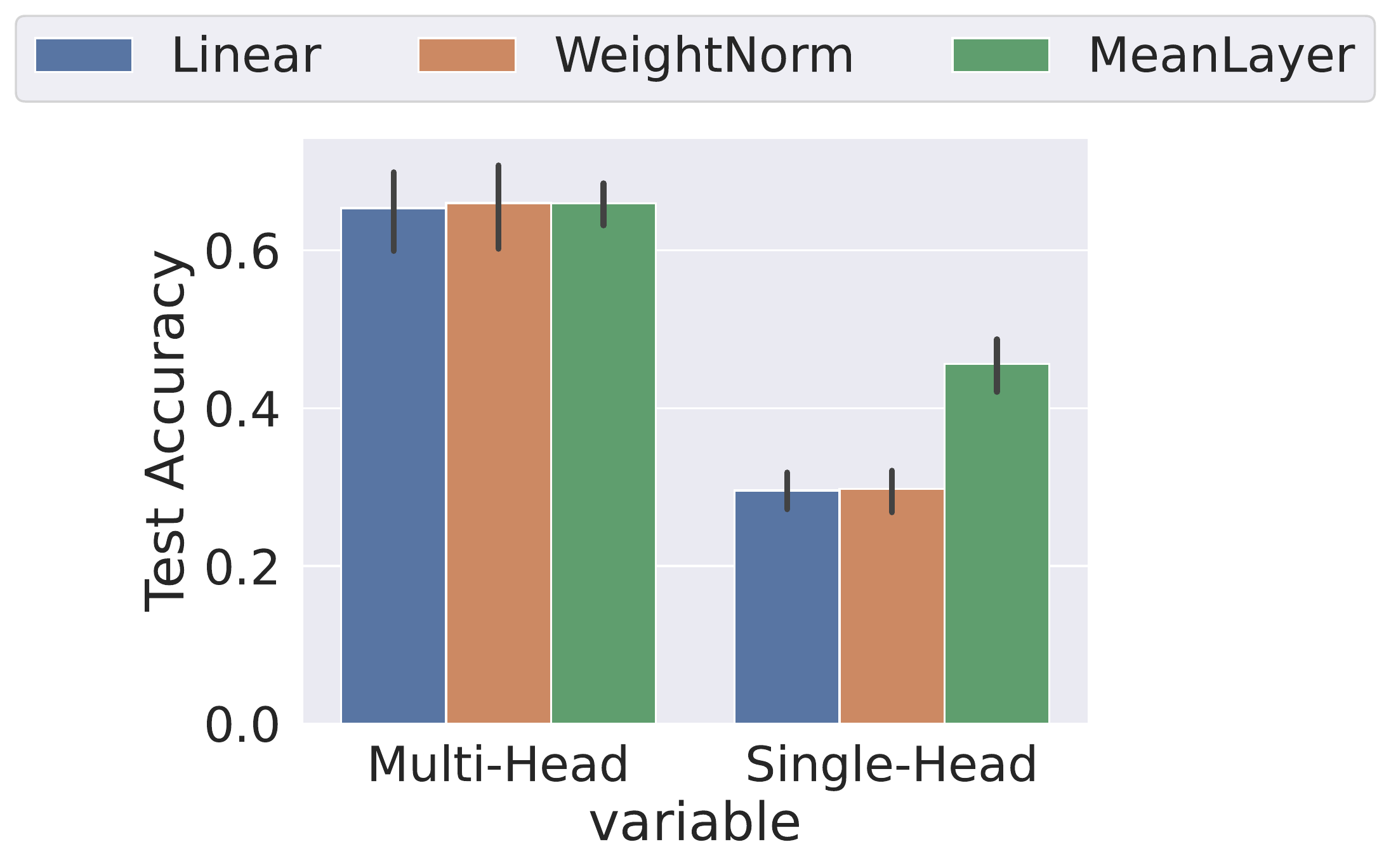}
    \includegraphics[width=0.24\linewidth,trim={0 1cm 0 0},clip]{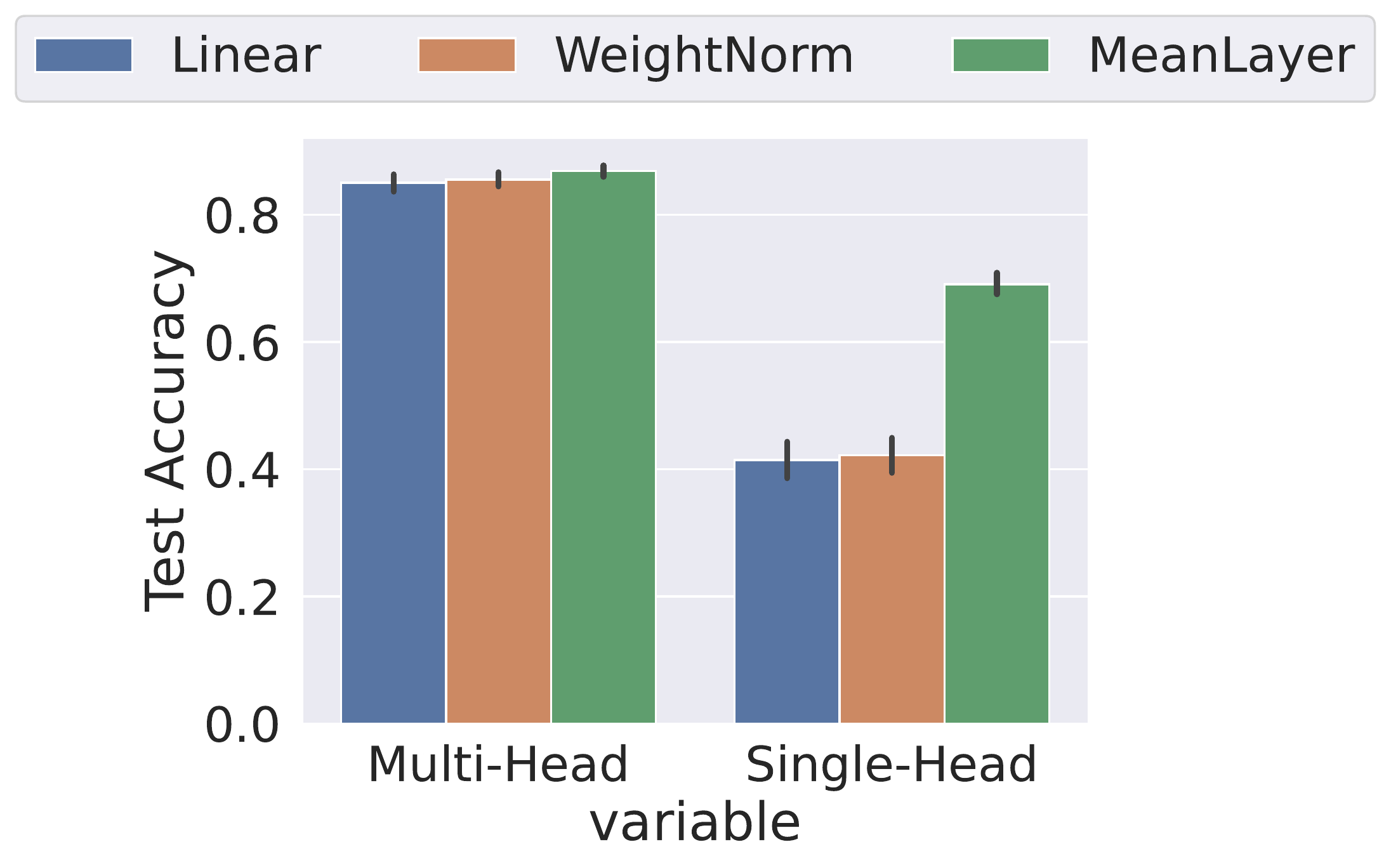}
    \includegraphics[width=0.24\linewidth,trim={0 1cm 0 0},clip]{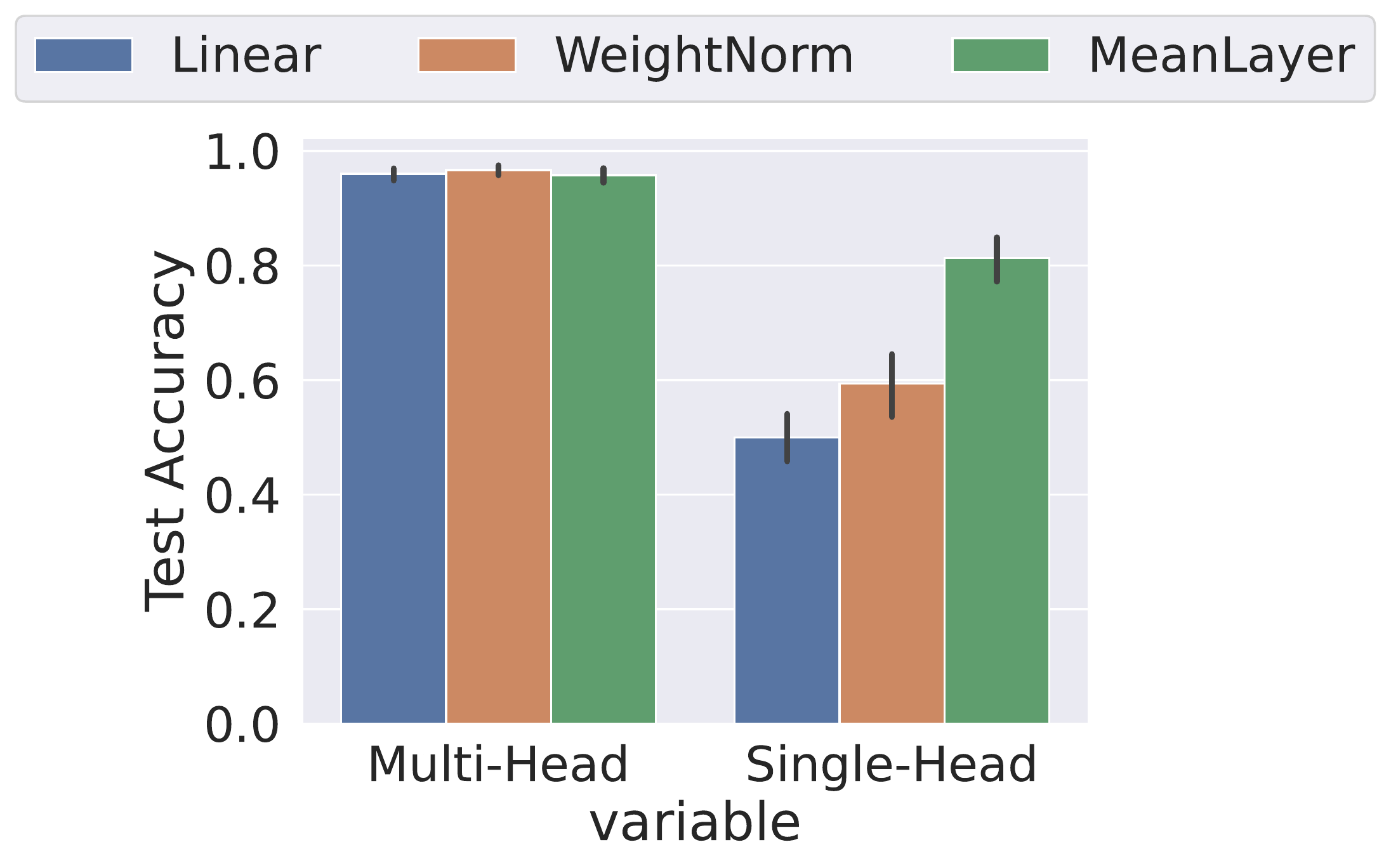}
    \caption{Comparison of test accuracy in multi-head and the same classifier without task label information (In this order: CIFAR10, CUB200, OxfordFlower102, OxfordPet). The performance gap in the linear layer indicates how local features selected are generalizable. We use the weightnorm to estimate if bias or norm imbalance phenomena \cite{lesort2021continual} play a role in the performance gap . The MeanLayer baseline indicates the difference of difficulty of both evaluations. The results of this figure indicate that classifiers rely on local spurious features to solve tasks.}
    \label{fig:MH}
\end{figure}

\subsection{Local Spurious Features Experiments}

\checked{We report results in Figure \ref{fig:MH}. To understand the figure, it is important to keep in mind that only the inference method changes between single head and multi-heads results. The classifier and the weights are the same (cf Sec. \ref{sub:local_setting}). 
The results for datasets experiments with several pre-trained models have been averaged. The details of the results are in appendix \ref{ap:local_details}.
We see that there is a considerable accuracy gap between multi-head $A_{te-local\_softmax}$ and single head performance in the linear layer $A_{te-global\_softmax}$.
Moreover, the gap for MeanLayer is significantly smaller than the gap for the linear layer. Hence, the difference in difficulty is not sufficient to explain the accuracy gap of the linear layer. 
And finally, the similar results for linear layer and weightnorm show that the norm/bias unbalance is not a problem that could explain the performance gap.
Therefore, as described in Sec \ref{sub:local_setting}, we can conclude that the classifier selected local spurious features, which leads to a poor generalization and a low single head performance.}

\checked{These phenomena prove that the drop in performance in class-incremental can be due to a bad feature selection and not necessarily to forgetting. This is a fundamental observation for future continual learning approaches because it means that the features learned and selected do not necessarily have to be kept in later tasks. While the continual learning bibliography focused mostly on forgetting or transfer, the spurious local features problem is also a great challenge to deal with to create efficient continual learning.}

\checked{Another interesting insight that this experiment gives is that contrary to experiments in section \ref{sub:solution} which show that using a pre-trained model can help for spurious features, it does not offer a solution to local spurious features. A good feature extractor is then not sufficient for a good feature selection by the classifier.}

\checked{Replay is an excellent solution to avoid local spurious features problems. Indeed, the replay process simulates an identically and independent distribution on all the tasks seen so far, which makes it possible to fix potential mistakes in past feature selections. Nevertheless, even if this solution already exists, it needs much supplementary computation when the number of tasks grows. Therefore, a better understanding of forgetting and feature selection in continual learning could enable more efficient methods.}

\section{Conclusion}

Continual learning algorithms are built to learn, accumulate and memorize knowledge through time to reuse them later. Memorizing unadapted features can have catastrophic repercussions on future performance.
Then, to learn general features, algorithms need to deal with spurious and local spurious features.

This paper investigates first the impact of spurious features on continual learning. 
Algorithms easily overfit spurious features for one or several tasks, leading to poor generalization. 
Our goal was to investigate if continual learning algorithms can benefit from the variation of the spurious features through time to ignore them.
Our results show that a classical continual learning approach such as rehearsal can deal with spurious features until a certain level of correlation with labels. However, even if this scenario is not very realistic, rehearsal is not sufficient once the spuriousness reaches 100\% of correlation with training labels, and alternative approaches should be found.

\checked{In a second round of experiments, we investigate the problem of local spurious features. Those features correlate spuriously with labels, not because of a covariate shift between train and test but because all data is not available at once. We show that algorithms can easily overfit local spurious features that are not generalizable. Consequently, the solution learned to solve a task becomes outdated when new data arrives and performance on past tasks decreases significantly.}

\checked{In the continual learning literature, performance decrease is generally attributed to catastrophic forgetting. Our results show that the problem of local spurious features seems to play also a major role.
More research is needed to understand better the impact of local spurious features along with catastrophic forgetting. Understanding this phenomenon is critical to better address forgetting and feature selection and enable efficient continual learning.}

\newpage

\section*{Acknowledgement}

We would like to thank Laurent Charlin, Pau Rodriguez, Massimo Caccia, Alexandre Ramé, Kartik Ajuha, Thang Doan, Oleksiy Ostapenko, and Michelle Liu for fruitful discussions and feedbacks on this work.

\bibliographystyle{abbrv}
\bibliography{continual,others}

\newpage
\appendix

\section{Compute}

The experiments were run on internal cluster on Quadro RTX 8000 GPUs. The total time of compute for an approximate period of 100 days of GPU use with hyper-parameters selections and experiments.

\section{Lowering the Support of the distribution at each task}
\label{ap:sec:support}

\subsubsection{Lowering the Support Amount}

\begin{figure}
    \centering
    \includegraphics[width=\linewidth]{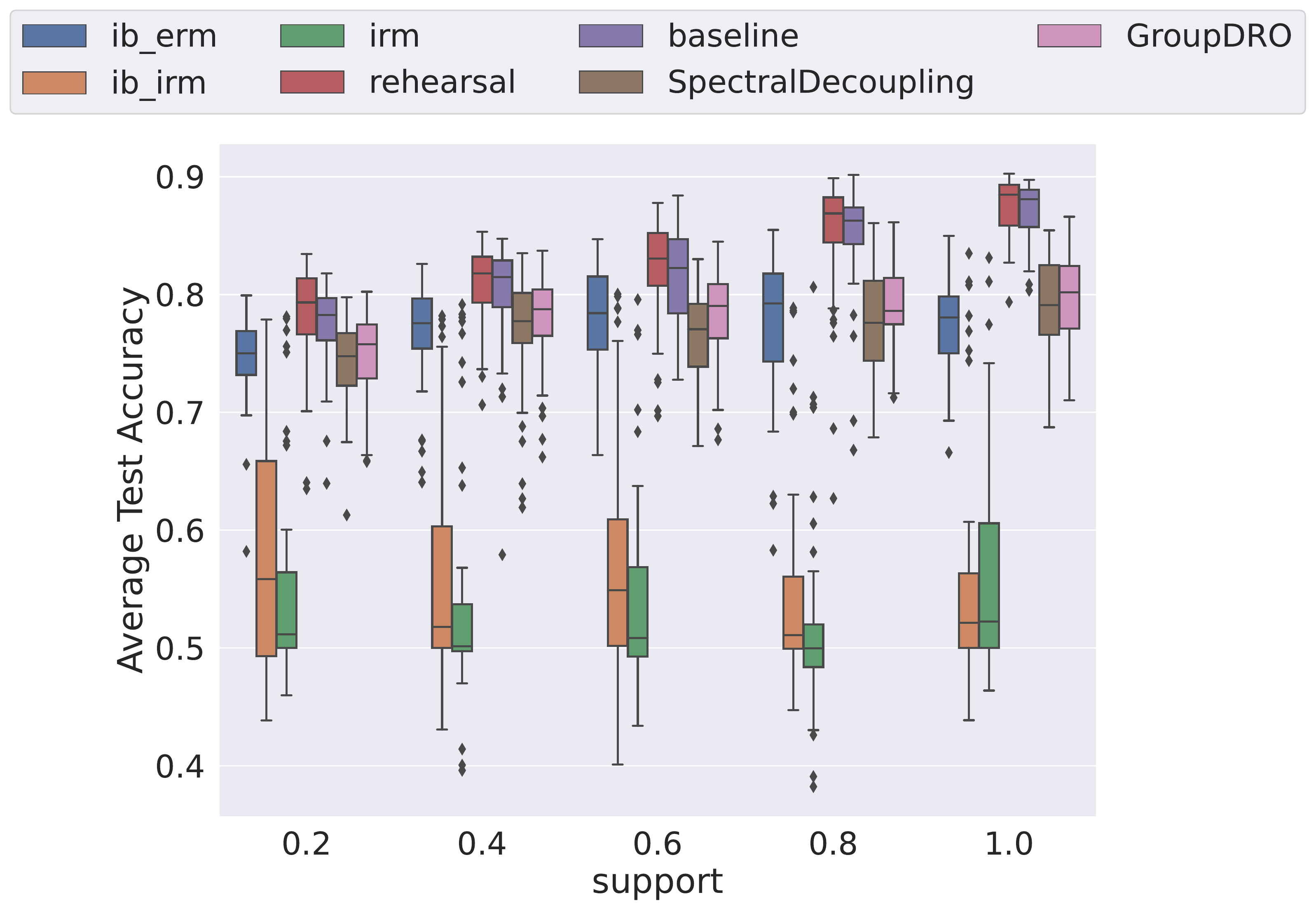}
    \caption{Averaged accuracy $\Omega$ on 10 tasks over various amount of support, the spurious correlation is set to $0.75$.}
    \label{fig:support_exp}
\end{figure}

In experiments Sec. \ref{subsub:correlation}, the same data support was used for all tasks. This means that the shared support of original data is 100\%. However, we expect continual algorithms to learn with less support. The support of a task is the percentage of the full original distribution within the task.
 We investigate in those experiments the influence that the support amount has on learning algorithms. We set the spurious correlation to $75\%$ in those experiments.

\checked{As it is designed in the SpuriousCIFAR2 scenario, each class (0 and 1) has 5 distributional modes corresponding to the 5 original classes. In the previous experiments, all the data for all those modes are available in all tasks only the spurious features changed from one task to another.}

\checked{In those experiments, we reduce the support by subsampling the original data. We select the data of a subset of the CIFAR10 classes for each task. For example, we select only cars for class 0 and only deers for class 1, instead of airplanes, cars, trucks, ships, and horses for class 0 and birds, cats, dogs, deers, and frogs for class 1.
Hence, if we use the support of $0.2$ (i.e. $20\%$), we will only select the data of $5*0.2=1.0$ original classes for class 0 and one other for class 1. For simplicity's sake, we will use only support compatible with the number of classes to have a round subset, i.e., $[0.2, 0.4, 0.6, 0.8, 1.0]$. At each task, the support is randomly sampled, hence the same original data can be in several tasks, but the spurious features will still be different for all tasks.}

The support experiments results, illustrated in Fig.~\ref{fig:support_exp}, show that 
in all support settings, the finetuning baseline and rehearsal are the best performing methods.
On the other hand, contrary to what would be expected, the amount of support seems to not play an important role in the final performance, at least in the range of supports possible in our scenario. 
This is probably because doing replay converts a partially observable setting into a fully observable setting by simulating an iid distribution.

\section{Discussion}
\label{ap:sec:discussion}

\textbf{Spurious Features vs Local Spurious Features: }
\checked{Spurious features and local spurious features lead to the same problem for learning algorithms: overfitting features that are not discriminative. The difference is that spurious features result from a covariate shift between training and test data. In contrast, local spurious features are due to the unavailability of all data while learning. Local spurious features are then, more specifically, a continual learning challenge, and we showed that they could have a significant impact in classical continual learning scenarios.}

\textbf{Solutions to spurious features: }
The problems of spurious features and local spurious features lead to phenomena where forgetting is helpful to improve final performance \cite{zhou2022fortuitous} also denoted as graceful forgetting. Indeed, forgetting by reinitializing some weights allows escaping spurious local minima and relearning a better solution taking new data and knowledge into account. 
On a more general note, spurious features and local spurious features make ineffective approaches that are too rigid and unable to modify and fix previously learned features/knowledge. The replay methods have been known to be a good solution for many continual learning problems. Our findings do not challenge this approach. However, it could be of some use to make replay approaches more effective and reduce the need for replay.

\textbf{Benchmark: }
The scenario CIFAR2Spurious proposed in this paper has been designed to highlight the problem that spurious correlation might create in continual learning.
This scenario plainly fulfills its task of disturbing CL algorithms, particularly in the $100\%$ correlation setting.
However, it can not be used as a benchmark to evaluate the robustness of algorithms. The spurious features are very simple, and a simple ad hoc processing of data could solve this scenario, i.e., encoding data with a pre-trained model as in Fig.~\ref{fig:pre-training}.
A proper benchmark to assess robustness to spurious correlation would propose spurious features easy to learn by the model but harder to detect or ignore than simple squares of color. DomainBed datasets \cite{gulrajani2020search} are an interesting set of benchmarks for spurious correlation investigation. However, the amount of features is not controllable, making it harder to evaluate the limits of algorithms.

\textbf{Potential negative societal impacts:} This work aims at providing insights into the impact that spurious correlation between features and labels can have on continual learning. The goal is to understand better the continual learning challenge for the later improvement of continual learning algorithms.
While continual learning could have some bad use-case, this work does not provide tools that push in this direction. 
We used only data about animals and objects for our experiments which we believe can not be used for questionable societal applications.

\section{Samples Support Experiments}
\label{ap:sec:samples}

\begin{figure}[ht]
    \centering
    \begin{subfigure}[b]{0.3\linewidth}
        \centering
        \includegraphics[width=\linewidth]{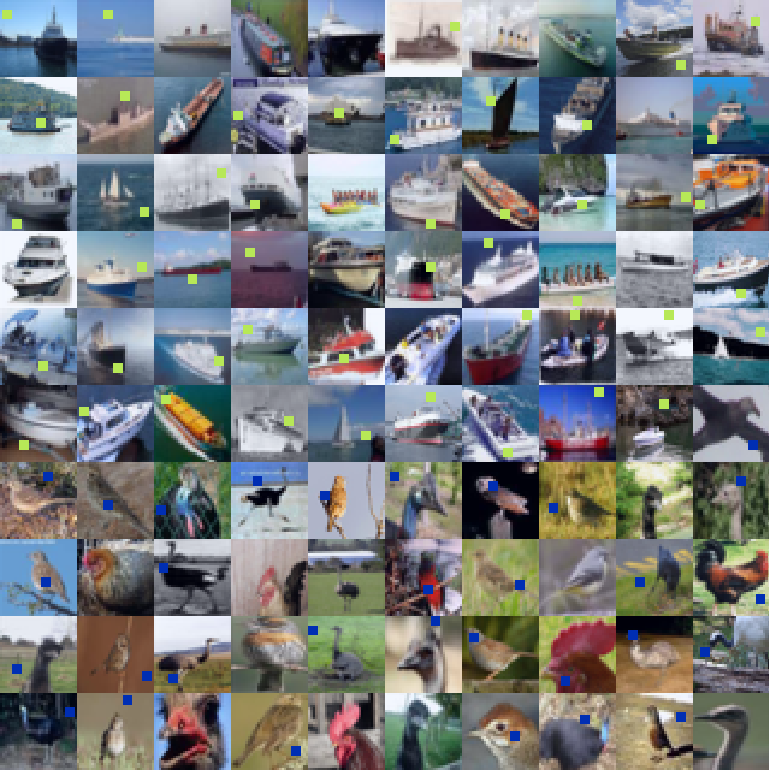}
    \caption{Train / Valid data task 0}
    \end{subfigure}
    \begin{subfigure}[b]{0.3\linewidth}
        \centering
        \includegraphics[width=\linewidth]{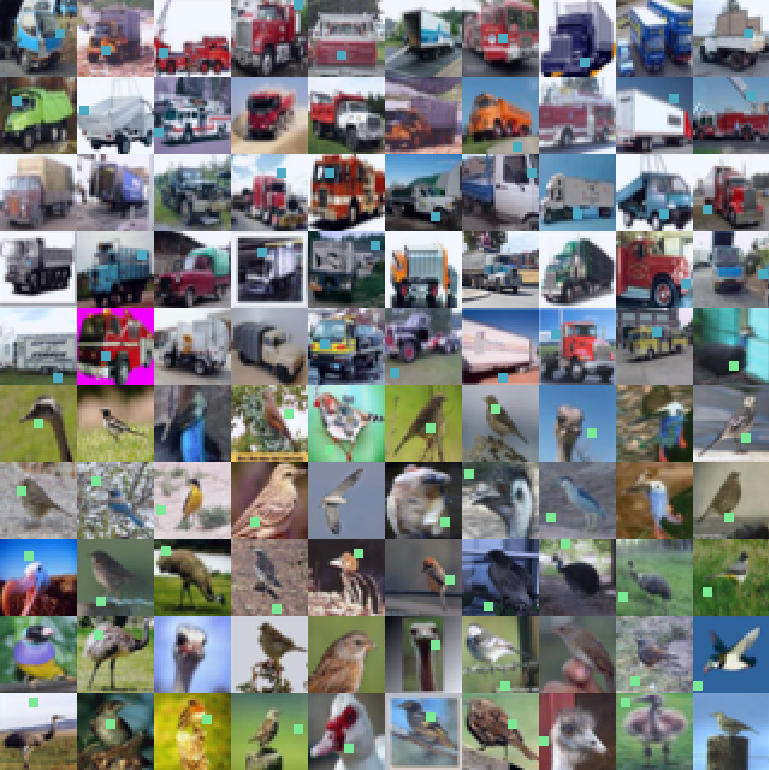}
    \caption{Train / Valid data task 1}
    \end{subfigure}
    \begin{subfigure}[b]{0.3\linewidth}
        \centering
        \includegraphics[width=\linewidth]{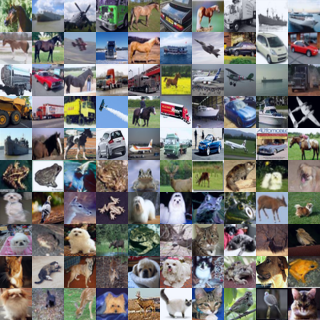}
    \caption{Final Test data}
    \end{subfigure}
    
    \caption{Samples for support experiences, here with 20\% support, i.e. the data of only two of the original classes in each task. }
    \label{fig:data_support}
\end{figure}

\section{Hyper-Parameters selection}
\label{ap:sec:HPs}

For a fair comparison between algorithms originally designed for continual learning, such as replay, and OOD algorithms, we conduct the hyper-parameter search more intensively for OOD approaches.

For each OOD approach, we search for the best hyper-parameters in the range proposed in the DomainBed GitHub repository \footnote{\textit{https://github.com/facebookresearch/DomainBed}}. We also searched for the best learning rate with the bayesian method of wandb \cite{wandb}. We used approximately 100 runs for each OOD baseline to select hyper-parameters. The scenario for hyper-parameters selection was an OOD setting with 5 environments of SpuriousCIFAR2 with $75\%$ correlation but all simultaneously available (with no continual stream of tasks).

The hyper-parameters for rehearsal and finetuning (baseline), have been selected on a finetuning training in a single task setting with $75\%$ of correlation. The number of samples per class for replay has been selected on a 5 task SpuriousCifar2 scenario with $75\%$ of correlation. 

\section{Bias Norm imbalance}
\label{ap:sec:bias_norm}

As described in \cite{lesort2021continual}:
 A linear layer is parameterized by a weight matrix $A$ and bias vector $b$, respectively of size $N \times h$ and $N$, where $h$ is the size of the latent vector (the activations of the penultimate layer) and $N$ is the number of classes.
For $z$ a latent vector, the output layer computes the operation $o = A z + b$.
We can formulate this operation for a single class $i$ with $\langle z, A_i \rangle + b_i = o_i$, where $\langle \cdot \rangle$ is the euclidean scalar product, $A_i$ is the $i$th row of the weight matrix viewed as a vector and $b_i$ is the corresponding scalar bias.

It can be rewritten:
\begin{equation}
\lVert z \rVert \lVert A_i \rVert \cdot cos(\angle(z, A_i)) + b_i = o_i
\label{eq:linear}
\end{equation}
Where $\angle(\cdot, \cdot)$ is the angle between two vectors and $\lVert \cdot \rVert$ denotes here the euclidean norm of a vector.

Then, at inference time, $y_i= argmax_i(o_i)$ rely on the norm of $\lVert A_i \rVert$ and on the bias $b_i$. 
Within a single task, i.e. within a single head in a multi-head setting, $\lVert A_i \rVert$ and $b_i$ are balanced to predict class correctly. However, we can not ensure that  $\lVert A_i \rVert$ and $b_i$ will are not biased from one head to another.

To avoid unbalance for bias and norm for inference, \cite{lesort2021continual} proposed the \textit{weigthnorm} layer where: $\lVert z \rVert \cdot cos(\angle(z_t, A_i)) = o_i$ and show that this layer in efficient in learning in incremental and lifelong settings.

\section{Details Multi-Head experiments}
\label{ap:local_details}

\begin{figure}[h]
    \centering
    
    \begin{subfigure}[b]{0.24\linewidth}
        \centering
        \includegraphics[width=\linewidth]{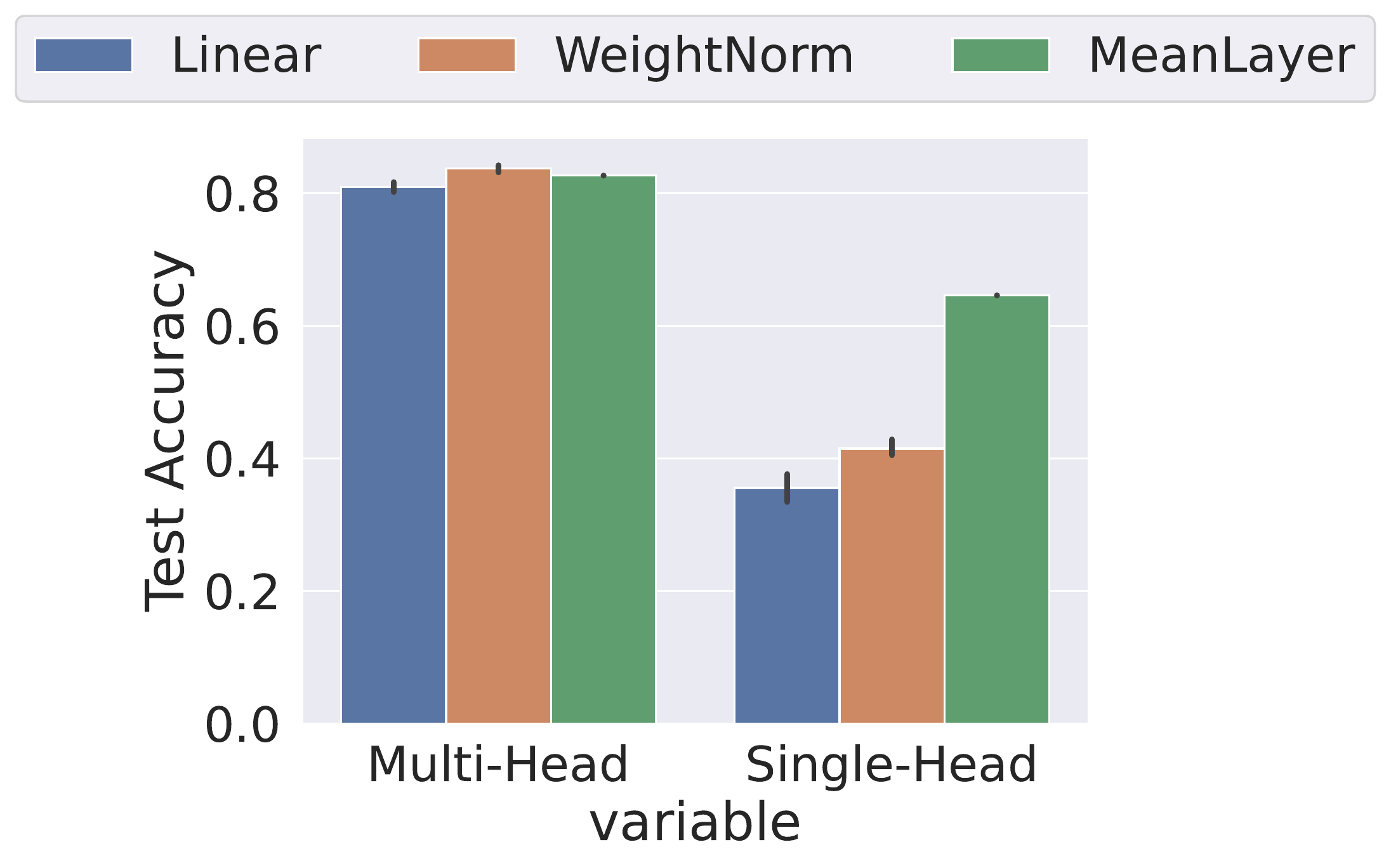}
        \caption{alexnet\\ OxfordFlower102}
    \end{subfigure}
    \begin{subfigure}[b]{0.24\linewidth}
        \centering
        \includegraphics[width=\linewidth]{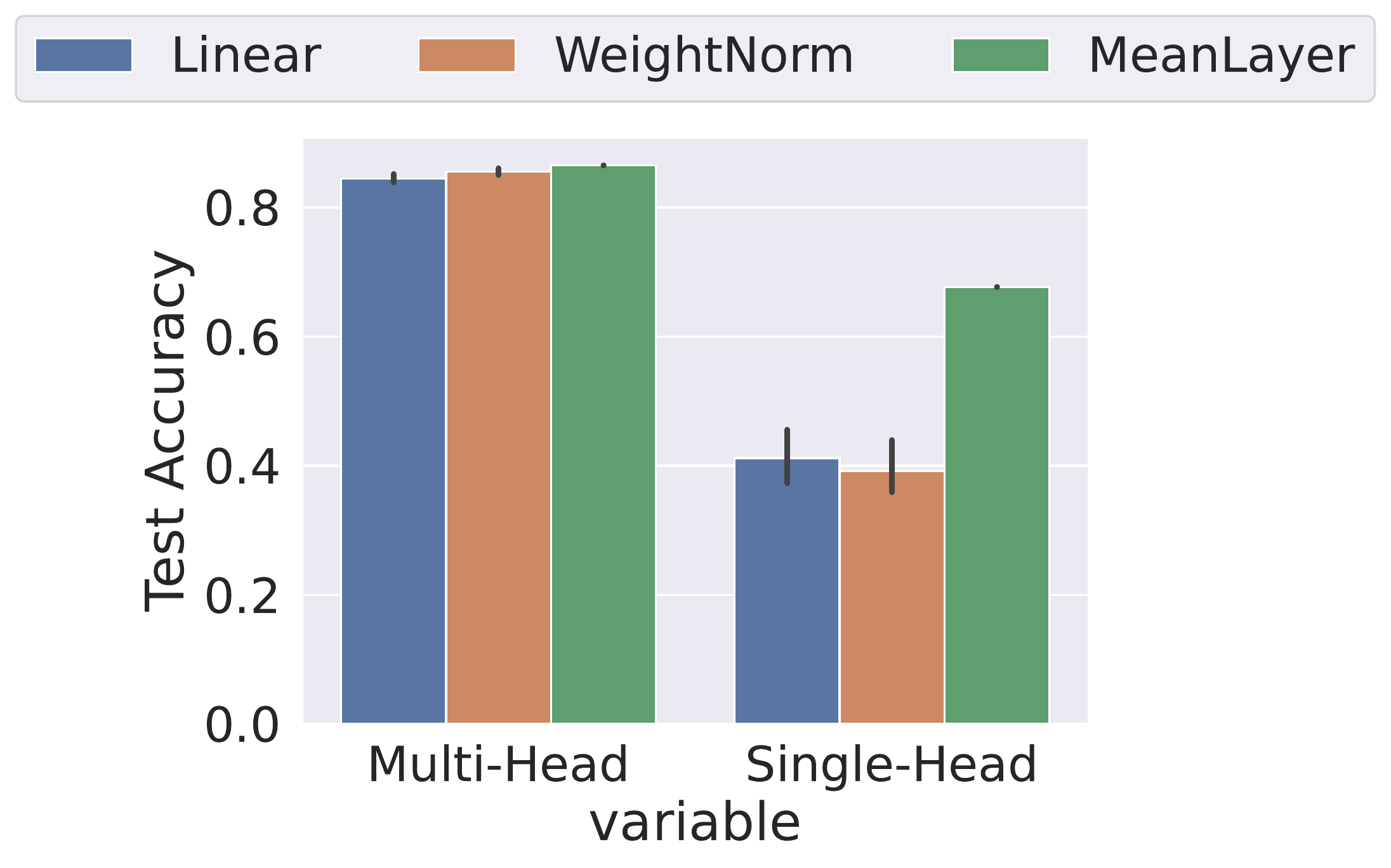}
        \caption{vgg\\ OxfordFlower102}
    \end{subfigure}
    \begin{subfigure}[b]{0.24\linewidth}
        \centering
        \includegraphics[width=\linewidth]{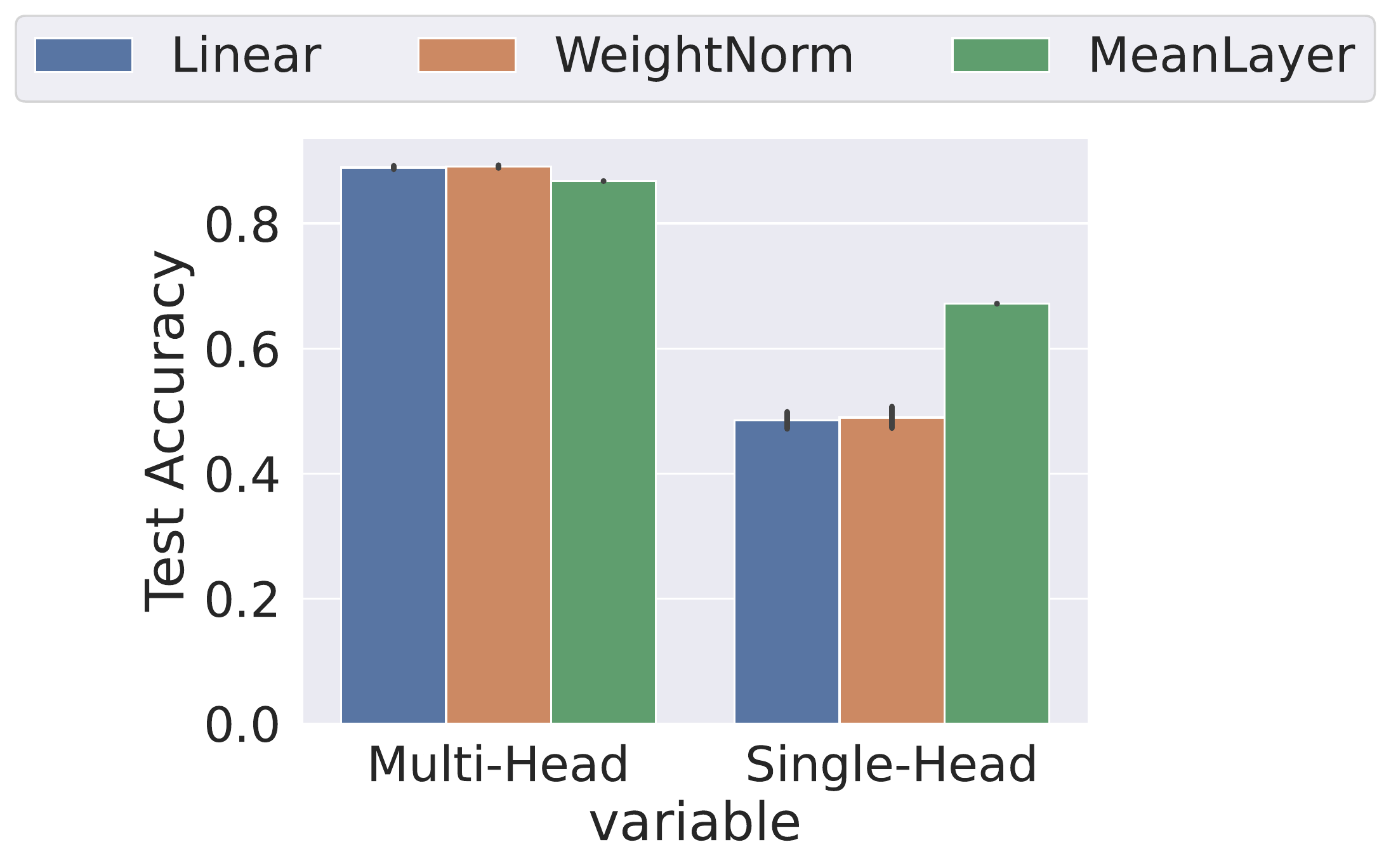}
        \caption{googlenet\\ OxfordFlower102}
    \end{subfigure}
    \begin{subfigure}[b]{0.24\linewidth}
        \centering
        \includegraphics[width=\linewidth]{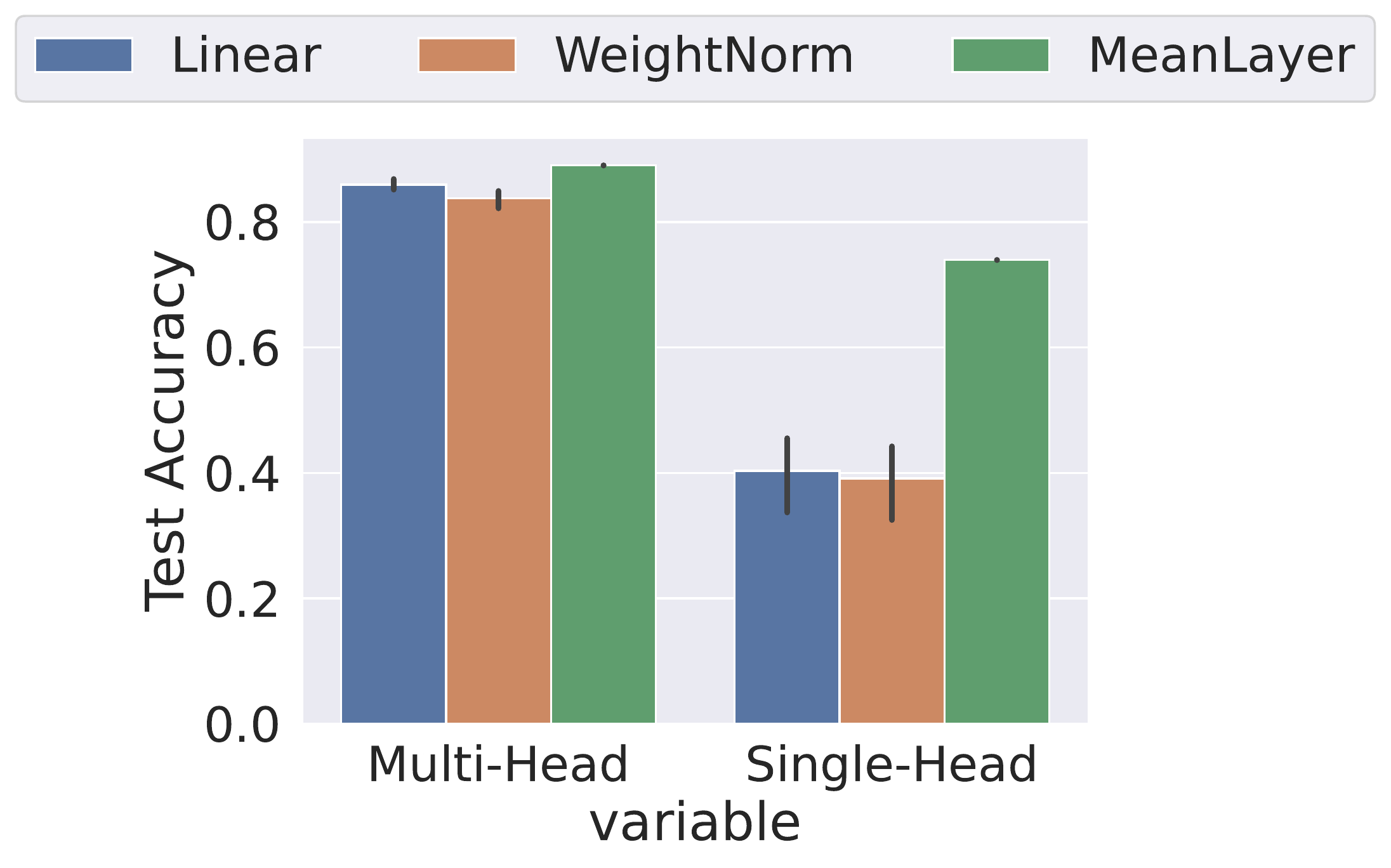}
        \caption{resnet\\ OxfordFlower102}
    \end{subfigure}

    \begin{subfigure}[b]{0.24\linewidth}
        \centering
        \includegraphics[width=\linewidth]{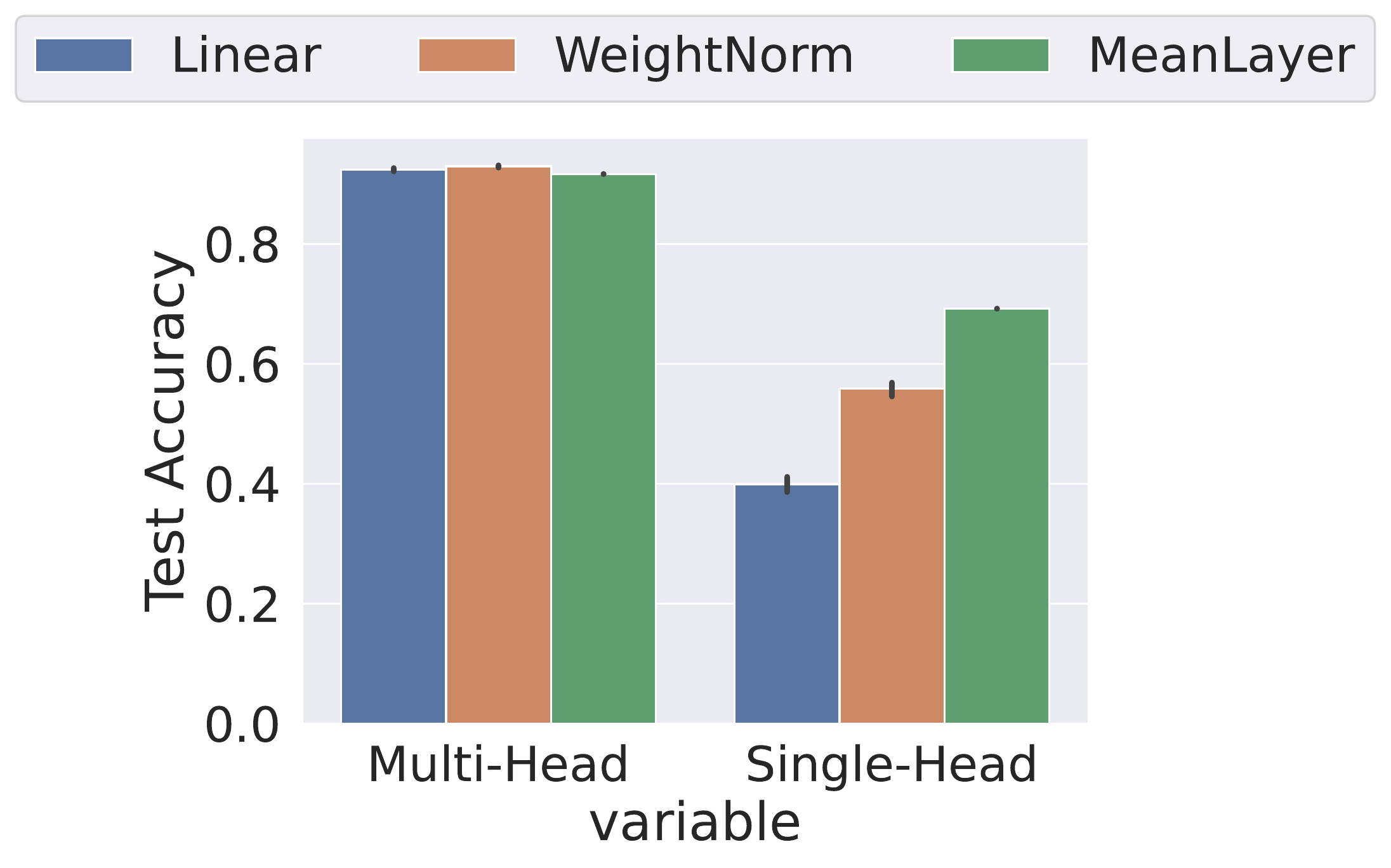}
        \caption{alexnet\\ OxfordFlower102}
    \end{subfigure}
    \begin{subfigure}[b]{0.24\linewidth}
        \centering
        \includegraphics[width=\linewidth]{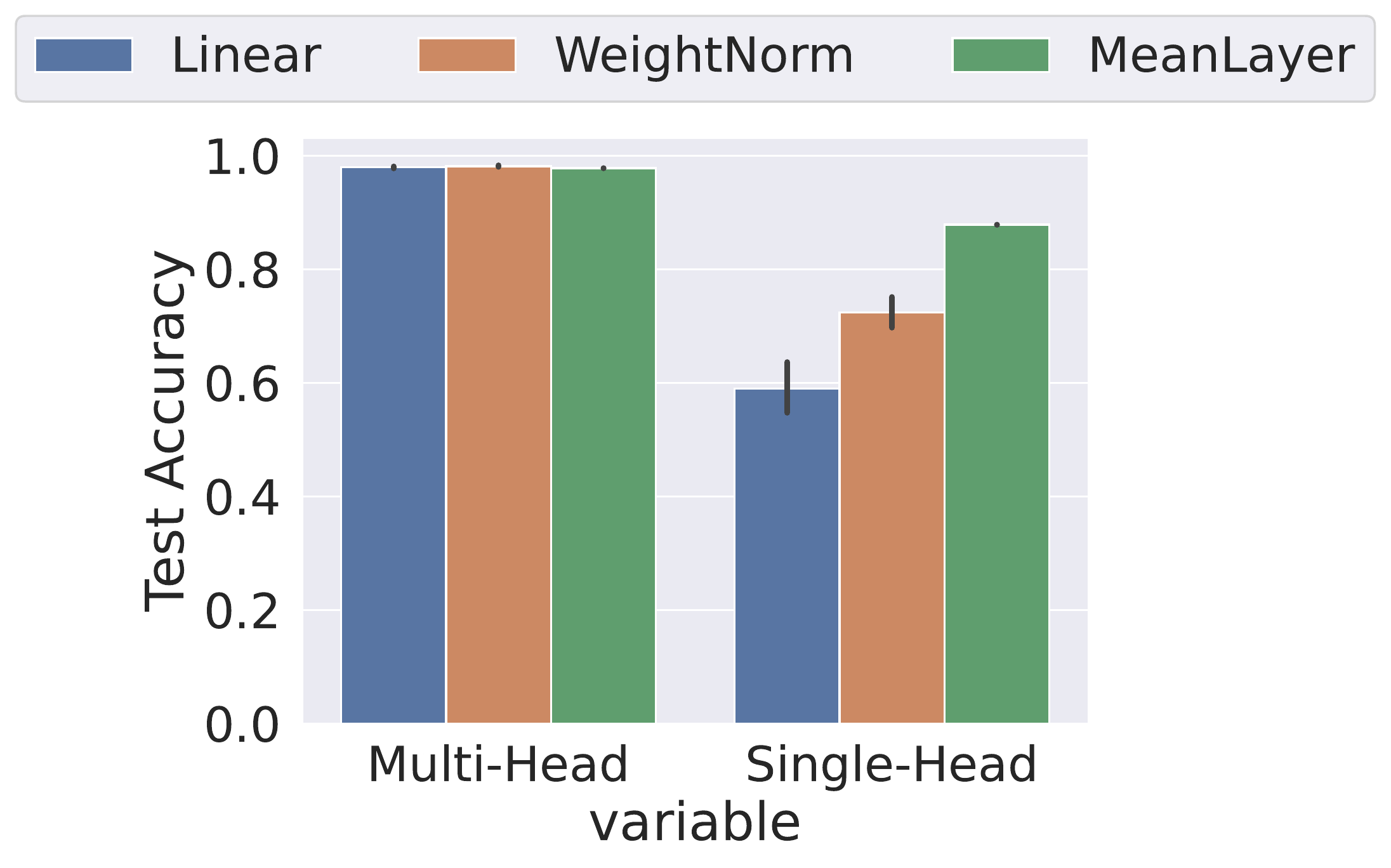}
        \caption{vgg\\ OxfordPet}
    \end{subfigure}
    \begin{subfigure}[b]{0.24\linewidth}
        \centering
        \includegraphics[width=\linewidth]{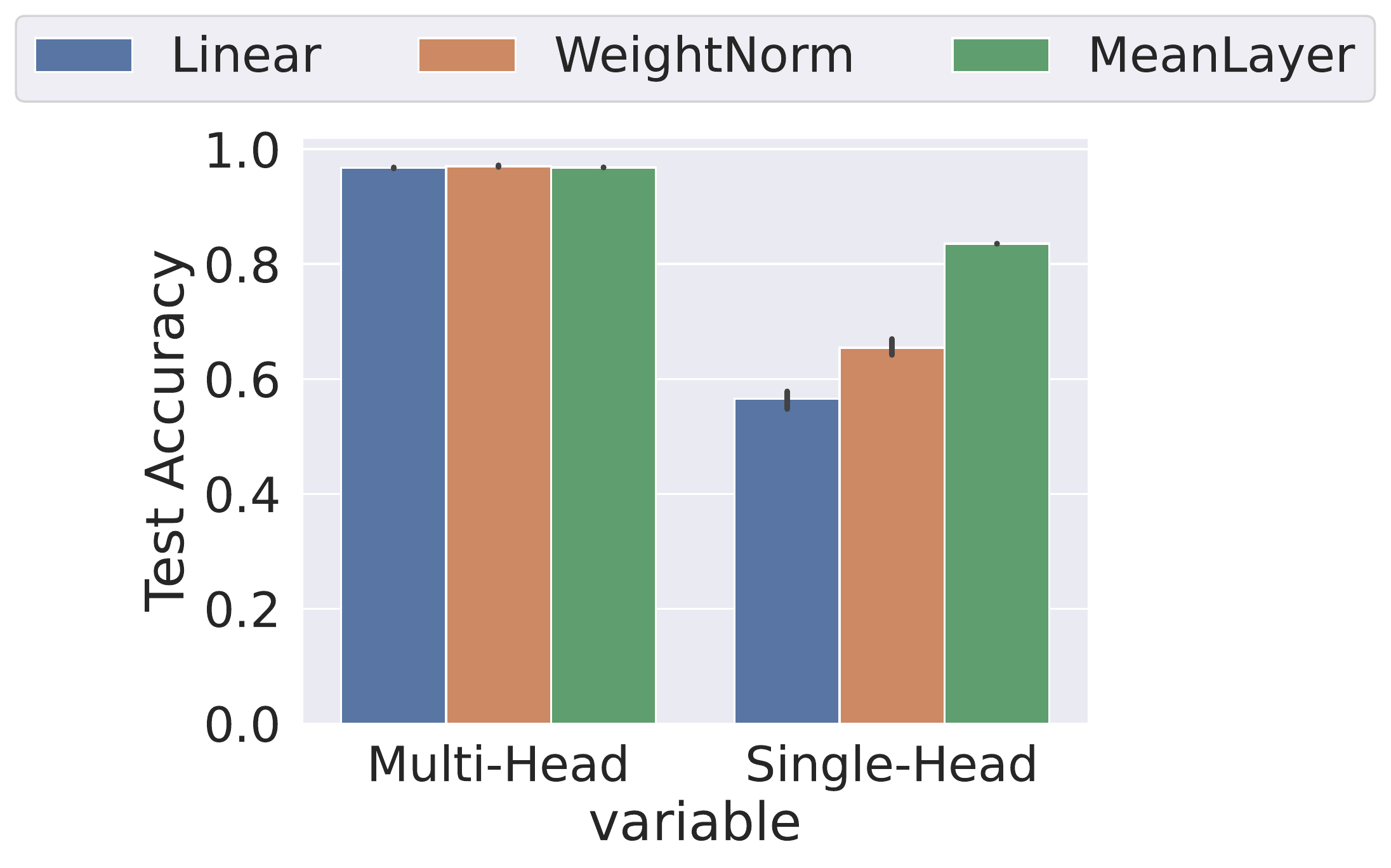}
        \caption{googlenet\\ OxfordPet}
    \end{subfigure}
    \begin{subfigure}[b]{0.24\linewidth}
        \centering
        \includegraphics[width=\linewidth]{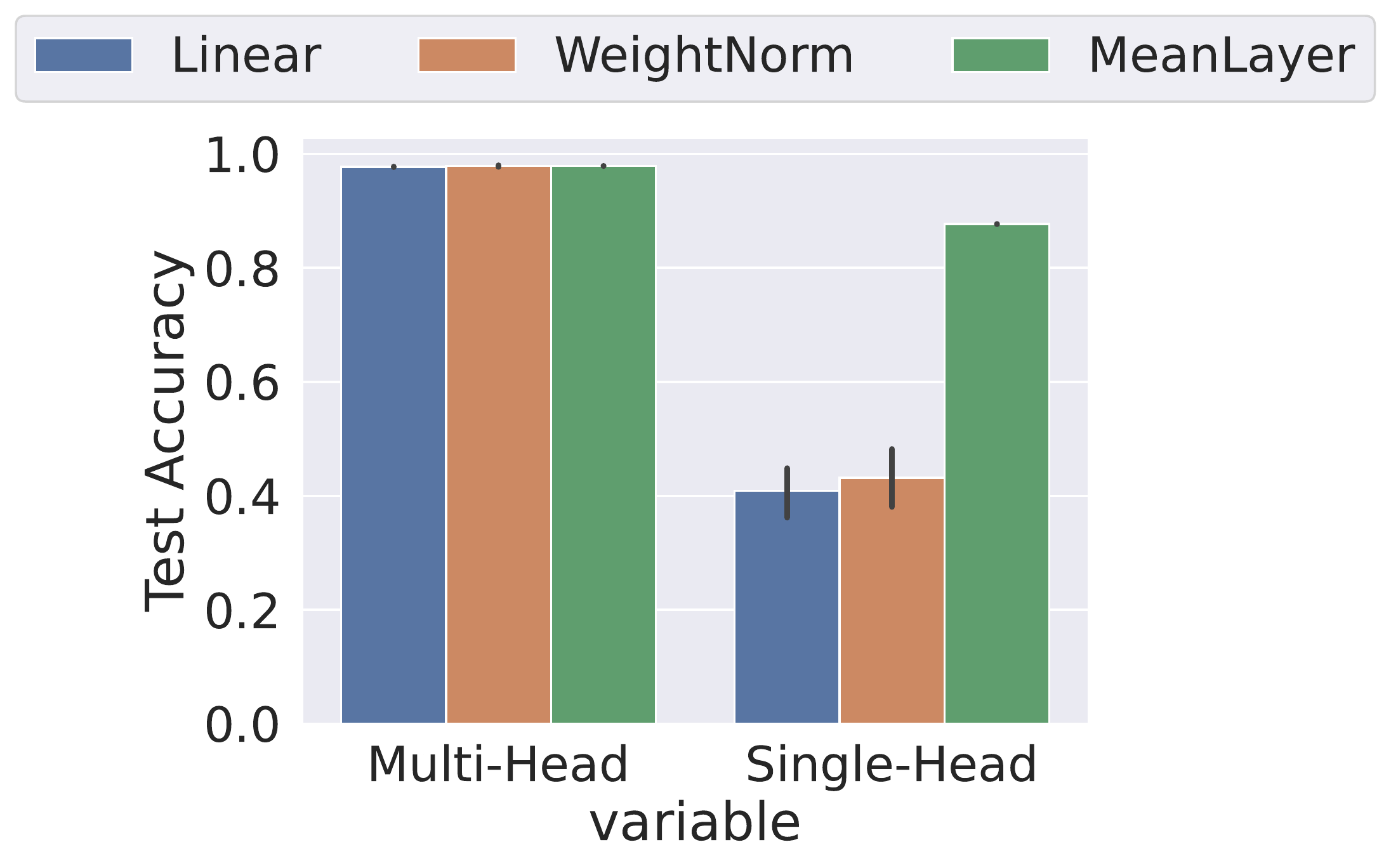}
        \caption{resnet\\ OxfordPet}
    \end{subfigure}

    \begin{subfigure}[b]{0.24\linewidth}
        \centering
        \includegraphics[width=\linewidth]{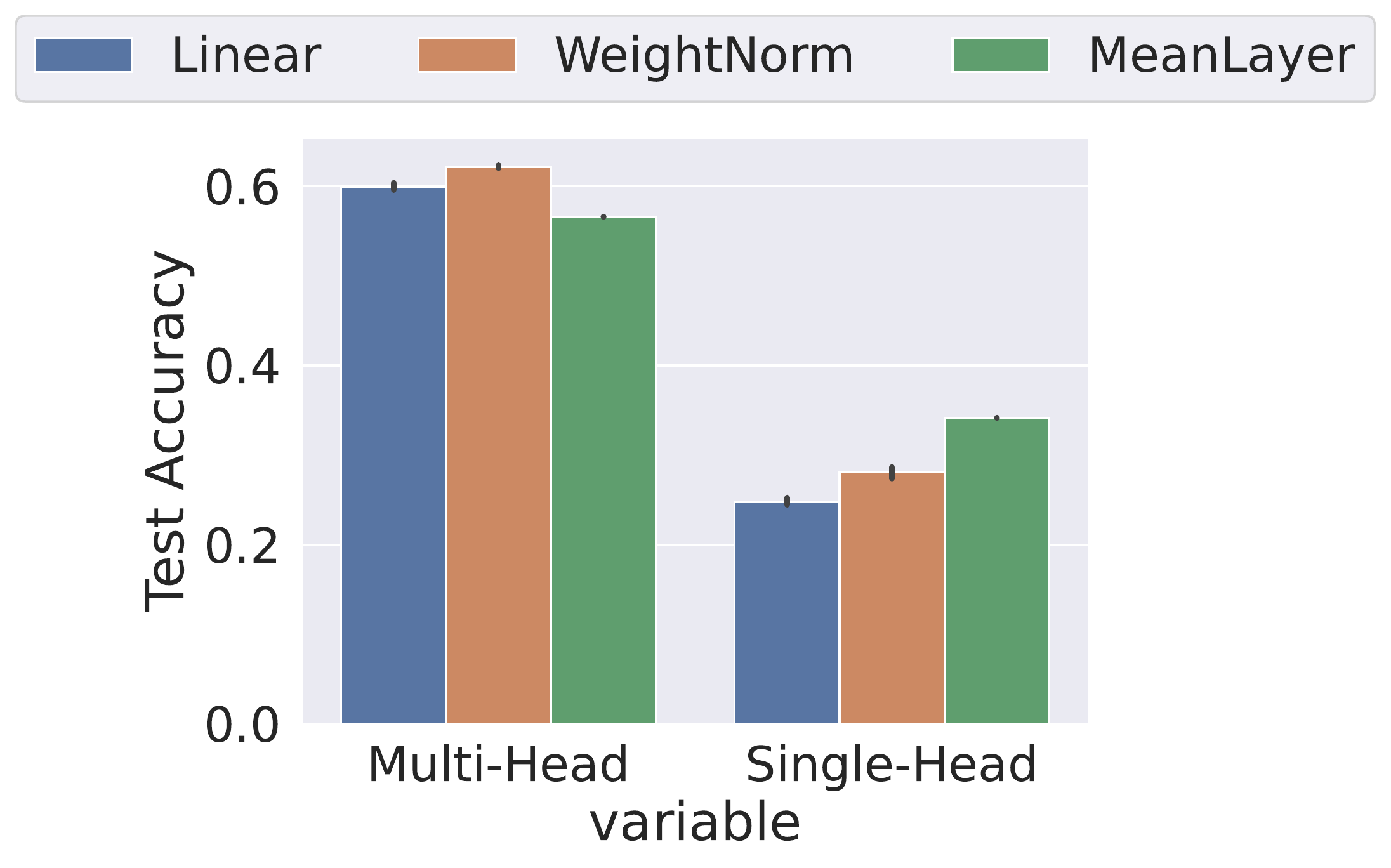}
        \caption{alexnet\\ CUB200}
    \end{subfigure}
    \begin{subfigure}[b]{0.24\linewidth}
        \centering
        \includegraphics[width=\linewidth]{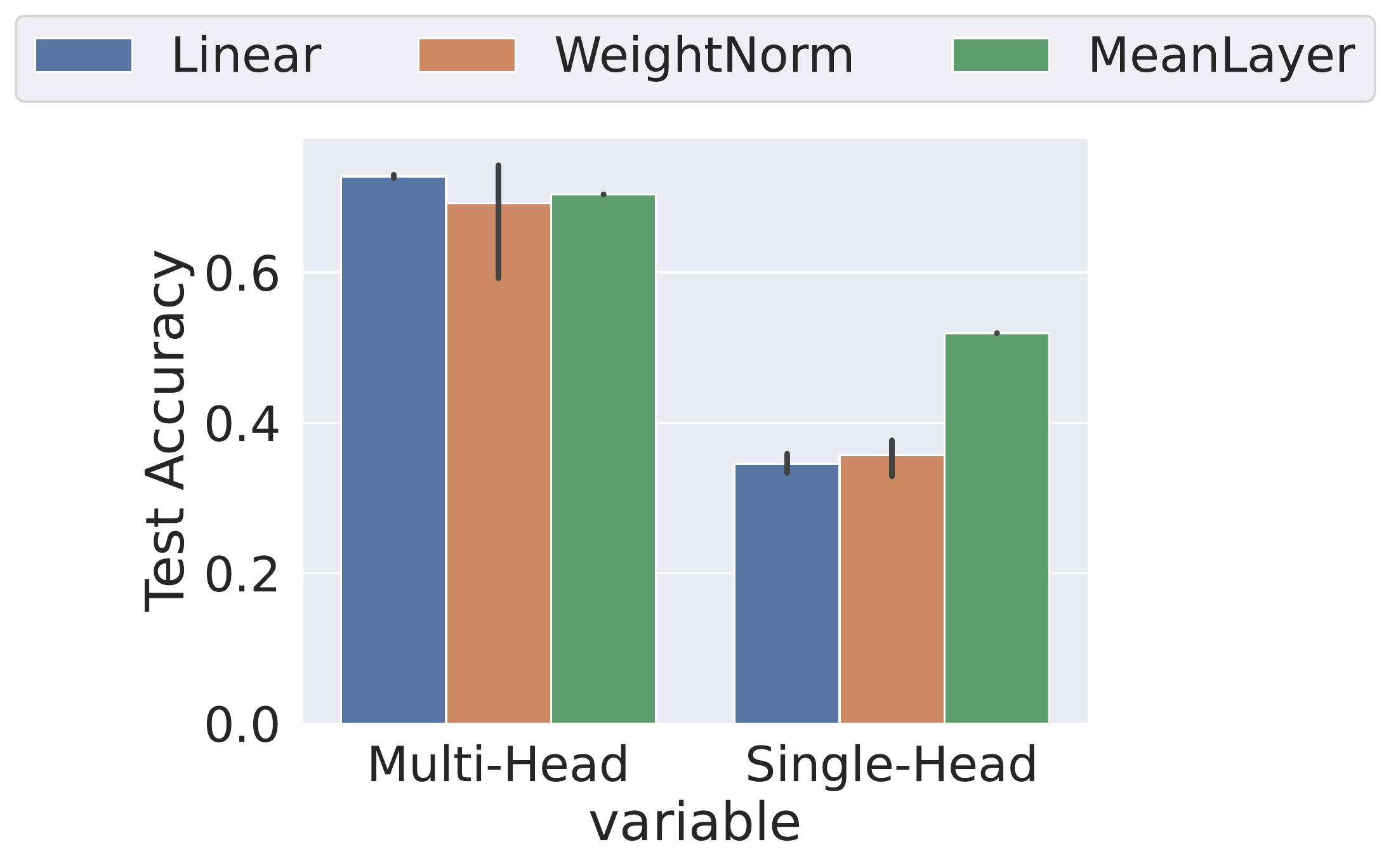}
        \caption{vgg\\ CUB200}
    \end{subfigure}
    \begin{subfigure}[b]{0.24\linewidth}
        \centering
        \includegraphics[width=\linewidth]{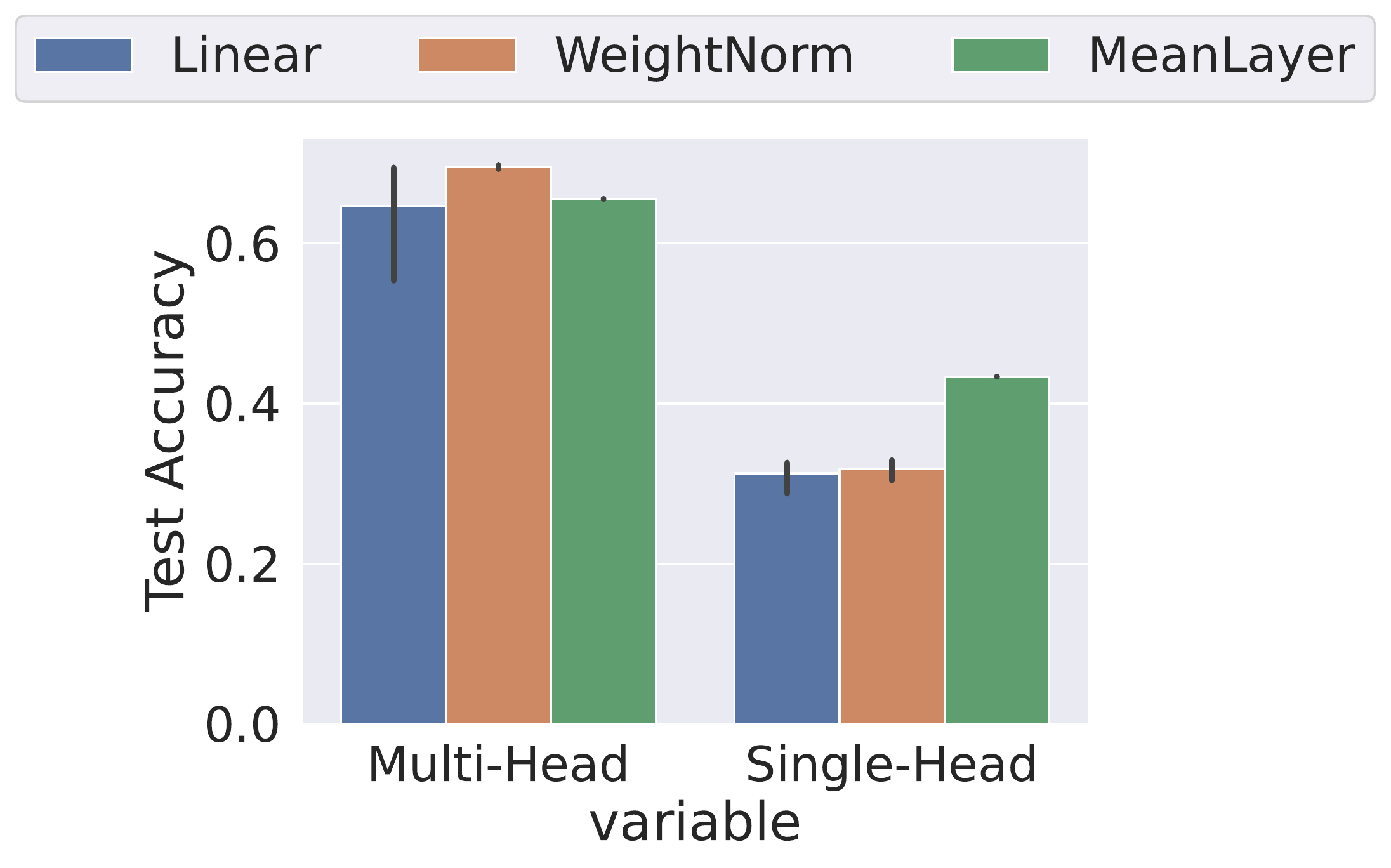}
        \caption{googlenet\\ CUB200}
    \end{subfigure}
    \begin{subfigure}[b]{0.24\linewidth}
        \centering
        \includegraphics[width=\linewidth]{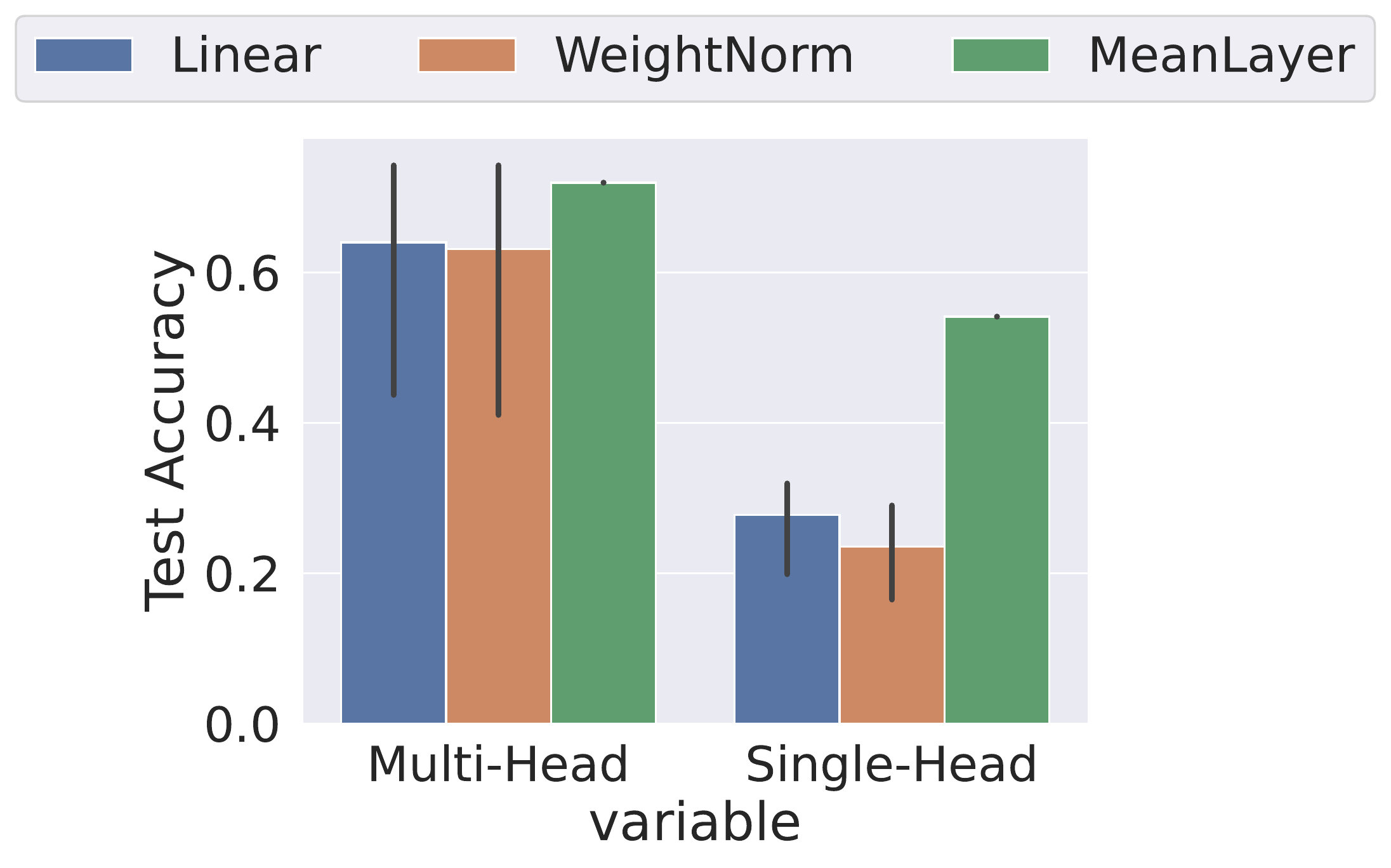}
        \caption{resnet\\ CUB200}
    \end{subfigure}
    \caption{Local spurious features experiments: after training in a multi-head way, we compare accuracy between multi-head (soft-max applied on a subset of classes' outputs) and single head (softmax applied on all classes' outputs). The performance differences assess how the model selected local spurious features to solve tasks. Two baselines are added, (1) meanlayer, which assesses the difference in the difficulty of the two evaluations, and (2) weightnorm, which assesses if the performance difference is due to an imbalance in norm or bias. The results in this figure show that models indeed rely on local spurious features to solve tasks.}
\end{figure}

\section{Consequences of Spurious features in CL Settings}
\label{ap:consequences}


\checked{In continual learning, two basic types of benchmarks are usually studied (1) Class-Incremental - new data brings new classes -, and Domain Incremental - new data are from already known classes but can be from another domain of the data distribution - \cite{lesort2021understanding,normandin2021sequoia}. In the presence of spurious features with covariate shift, the spurious features are in the training \textbf{and} the validation data of each task but are not in the final test set. }

\subsection{Class-Incremental}

\checked{Class-Incremental settings are designed to evaluate the capacity of models to learn incrementally to classify classes. By definition and for evaluation purposes, classes are seen in only one task and never again.
This is a good way to evaluate the capacity to learn, retain information, and avoid catastrophic forgetting. Still, it does not make it possible to reevaluate past knowledge with new data \cite{caccia2020online}.}

\checked{The only way to improve the features learned in past tasks/classes is then to do replay to confront past data with new data \cite{lesort2019regularization}.
In this setting, learning on the data of one class should lead to learning features that are good for in-distribution data (past and current classes) and out-of-distribution data (future classes).
The future data being unavailable, it is highly probable that some of the learned features locally are not global discriminative features and are, consequently, local spurious features.
Hence, the risk of spurious features is not negligible in class-incremental scenarios.}

\checked{We can note that catastrophic forgetting is often assimilated to important weights modification \cite{farajtabar2019orthogonal,Doan2021Theoretical}; however, spurious correlation (and even local features) can also decrease performance without modifying important weights. Models that need to learn new features to understand past tasks better are forced to operate a representation drift \cite{caccia2021reducing} (also called projection drift \cite{lesort2021continual}) to maximize final accuracy.}

\checked{Hence, this setting could be used to analyze local vs. global vs. spurious features. However, there is a risk that catastrophic forgetting interferes with bad features and entangles results. We propose first using domain incremental setting, where catastrophic forgetting is generally less intense, to simplify the problem and better highlight spurious features' troubles. In a second time, we will use a Class-Incremental scenario to highlight local spurious features challenges.}

\subsection{Domain Incremental}

\checked{Domain incremental settings are scenarios where the classes stay the same, but the data distribution changes. This setting is a way to evaluate how models can improve their knowledge/understanding of a concept or a class. The goal of the models is to find invariant features from several domains to predict accurately on all domains. It can also accumulate features that are complementary to characterize a class but that are not necessarily invariant.}

\checked{In continual learning, famous domain incremental settings are permutMNIST and rotated MNIST. Still, recent research in out of distribution generalization proposes interesting benchmarks that can be adapted into a domain incremental setting, such as in DomainBed datasets \cite{gulrajani2020search}.}

\checked{In this paper, we would like to study how approaches can deal with spurious correlation in a continual learning setting. Ideally, even if the model can not perfectly learn with spurious features in the first task, we would like that after seeing many domains (or environments), the model learns how to ignore spurious features and find invariant features if possible.}

\checked{This paper will experiment mainly with domain incremental since this setting is easier and compatible with most approaches. Moreover, we also experiment with simple class-incremental experiments to show that spurious local features have a significant impact in this setting.}

\section{Sampling Algorithm}
\label{ap:sec:algo_sampling}

\checked{Algorithm \ref{alg:train_sampling} describes the sampler used to train the model with replay. The input dataset $\dataset$ is a concatenation of the buffer with the current data. The goal is to make the probability of sampling on each class, the same whatever the number of sample for each class in $\dataset$.}

\begin{algorithm}[!t]
\caption{{\bf Balanced Sampling of Data Mixture.}}
\begin{algorithmic}[1]
  \Procedure{get\_sampler}{$\dataset$}
    \State $y \gets \dataset.y$ \Comment{{\tiny Get data class labels}}
    \State nb\_per\_class = \texttt{bincount}(y) \Comment{{\tiny count the number of occurence of each class}}
    \State $weights\_per\_class = \frac{1}{nb\_per\_class}$
    \State sample\_weights = weights\_per\_class[y]  \Comment{{\tiny give a sample probability weight to each data point}}
    \State sampler = \texttt{Sampler}($\dataset$, sample\_weights, replacement=True) \Comment{{\tiny create sampler to sample accordingly to the sample\_weights}}
    \State \textbf{return} sampler
  \EndProcedure
\end{algorithmic}
\label{alg:train_sampling}
\end{algorithm}

\end{document}